\documentclass[accepted]{styles} 

\usepackage[american]{babel}

\usepackage{natbib} 
\usepackage{mathtools} 
\usepackage{booktabs} 
\usepackage{tikz} 

\usepackage{amsmath}
\usepackage{amssymb}
\usepackage{amsthm}
\usepackage{amsfonts}
\usepackage{mathrsfs}
\usepackage{upgreek}
\usepackage{xcolor}
\usepackage{enumitem}
\usepackage{url}
\usepackage{graphicx}
\usepackage{mathtools}
\usepackage[normalem]{ulem}
\usepackage{multirow}

\newcommand{\BOUND}{{\mathcal{L}}}
\newcommand{\D}{{ \mathbb{D} }}

\newcommand{\E}{{\mathbb{E}}}
\newcommand{\SUMFLOWS}{{\sum_{b=1}^{B}}}

\newcommand{\f}{\mathbf{f}_\lambda}

\newcommand{\invf}{\text{inv}\mathbf{f}_\lambda}

\newcommand{\done}[1]{{\textcolor{violet}{\textbf{DONE:} #1}}}
\newcommand{\remark}[1]{{\textcolor{red}{[#1]}}}
\newcommand{\comment}[1]{\remark{#1}}

\renewcommand{\rho}{\pi}
\def\code#1{\texttt{#1}}

\usepackage[normalem]{ulem}

\newcommand{\old}[1]{\textcolor{red}{\sout{#1}}}

\renewcommand{\done}[1]{}
\renewcommand{\comment}[1]{}

\renewcommand{\old}[1]{}

\title{Reliable Categorical Variational Inference \\with Mixture of Discrete Normalizing Flows}

\author[1]{\href{mailto:Tomasz Kusmierczyk <tomasz.kusmierczyk@gmail.com>?Subject=Your MDNF paper}{Tomasz Ku\'smierczyk}{}} 
\author[2]{\href{mailto:Arto Klami <arto.klami@helsinki.fi>?Subject=Your MDNF paper}{Arto Klami}{}}
\affil[1,2]{%
    Helsinki Institute for Information Technology HIIT\\
    Department of Computer Science, University of Helsinki
}

\begin{document}
\maketitle

\begin{abstract}
Variational approximations are increasingly based on gradient-based optimization of expectations estimated by sampling. Handling discrete latent variables is then challenging because the sampling process is not differentiable.
Continuous relaxations, such as the Gumbel-Softmax for categorical distribution, enable gradient-based optimization, but do not define a valid probability mass for discrete observations. In practice, selecting the amount of relaxation is difficult and one needs to optimize an objective that does not align with the desired one, causing problems especially with models having strong meaningful priors. We provide an alternative differentiable reparameterization for categorical distribution by composing it as a mixture of discrete normalizing flows. It defines a proper discrete distribution, allows directly optimizing the evidence lower bound, and is less sensitive to the hyperparameter controlling relaxation. 
\end{abstract}

\section{INTRODUCTION}
\label{sec:introduction}

Efficient gradient-based algorithms for variational inference (VI) are nowadays routinely used for \emph{model-independent} Bayesian inference.
VI learns an approximation $q(x) \approx p(x|\D)$ of some latent variables $x$ conditional on observed data $\D$.
We express the distribution $q$ using a differentiable \emph{reparameterization} $\f(u)$, so that it is characterized by a base distribution $p_u(u)$ and a non-linear transformation parameterized by $\lambda$.
The parameters $\lambda$ are optimized
with stochastic gradients of 
the evidence lower bound $\BOUND$ (ELBO) 
\begin{align}
\nabla \BOUND  &= \nabla \mathbb{E}_{q_x(x)} \left[
\log \frac{p(\D, x)}{q_x(x)}
\right] \label{eq:elbo} \\
  &\approx 
  \frac{1}{S} \sum_{u \sim q_u(u)} \nabla\left[  \log \left( p(\D, \f(u))  \right) - \log q_x(\f(u)) \right], \notag
\end{align}
but other variational objectives could also be used.

Handling discrete variables within this framework is 
challenging as the gradient does not exist. 
However, 
most discrete distributions can be expressed in terms of continuous random variables by suitable auxiliary variable augmentation, for example, the binomial and negative binomial distributions using the Polya-Gamma distribution \citep{Polson13}, and the categorical distribution using the Gumbel distribution \citep{Gumbel54}. For categorical distribution this reparameterization is not yet sufficient for gradient-based learning because the required argmax in
$
x = \arg\max_j \left(
\log p_j + g_j
\right),$
where $
g_j \sim \text{Gumbel}(0,1)
$,
is not differentiable. For learning we hence need to use an \emph{approximation} replacing the argmax with a continuous \emph{relaxation} in form of the softmax function. Both \citet{maddison2017concrete} and \citet{jang2017categorical} presented simultaneously near-identical treatment of a relaxation scheme (called \emph{Gumbel-Softmax~(GS) relaxation} in this work) for categorical distribution, today routinely used in various learning tasks. 

GS relaxation 
has serious limitations as a variational approximation. It does not define a valid discrete distribution $q(x)$, preventing evaluation of the entropy, and hence one needs to optimize adjusted approximate objectives. 
Often these objectives do not match well with the true one, and the relaxation is highly sensitive to its hyperparameters with no obvious criteria for selecting them in pure inference tasks. These issues have remained somewhat undisclosed in the literature thanks to highly flexible decoders used in VAEs. 
However, for models with fixed likelihoods and strong priors GS relaxation is unreliable and burdensome. 

\begin{figure}[t]
\begin{center}
\includegraphics[width=0.9\columnwidth]{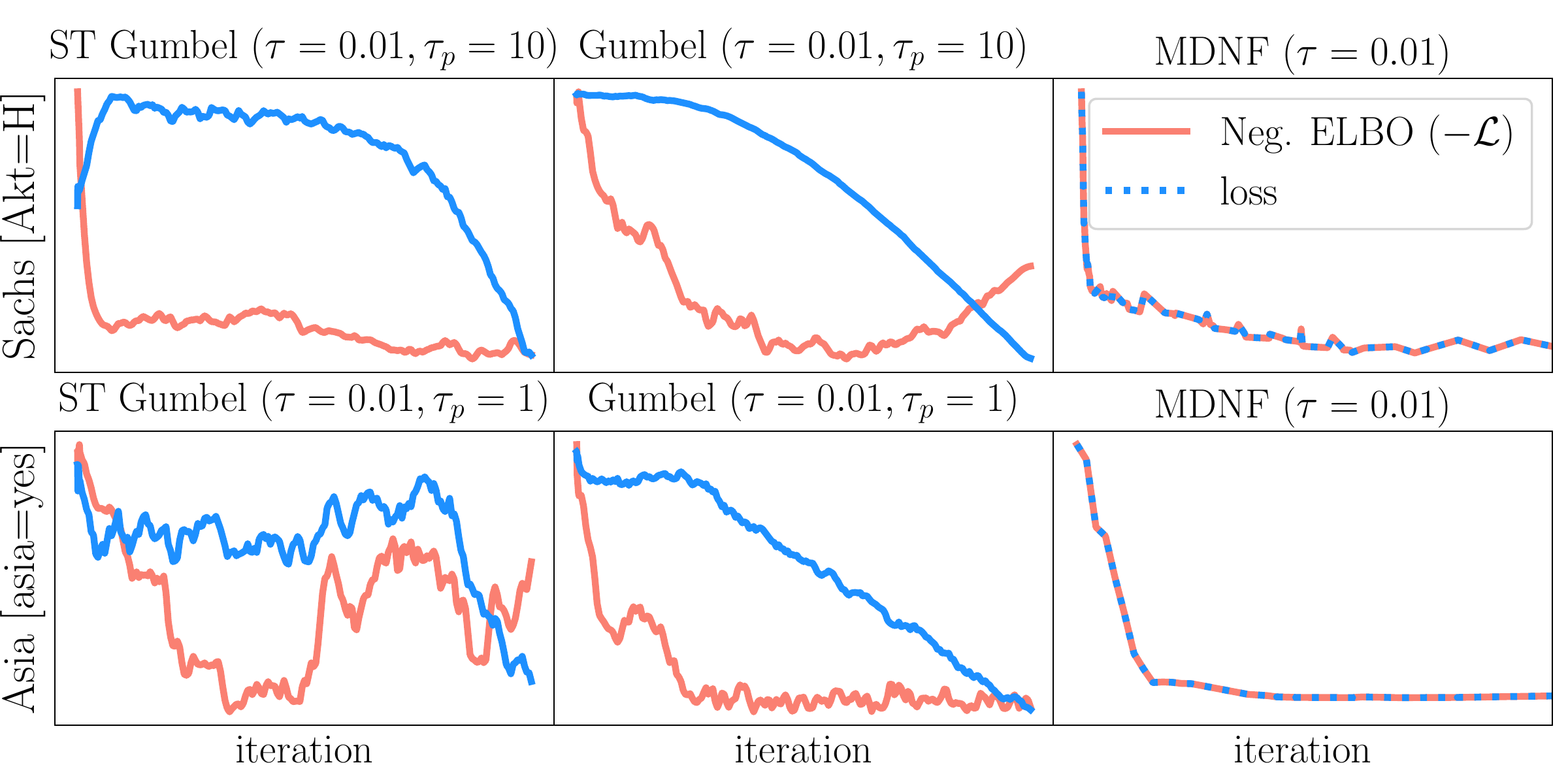}
\end{center}
\caption{
Minimization of Gumbel-Softmax loss (blue; with or without ST) does not necessarily minimize the objective (red) for the true discrete model.
The best model can be picked only based on  external validation, whereas the proposed MDNF minimizes the desired objective directly. See Section~\ref{sec:hyperparameter_experiment} for details.
}
\label{fig:bn_convergence}
\end{figure}

We propose a new differentiable parameterization for categorical distributions that builds on discrete normalizing flows (DNF) \citep{tran2019discrete}. DNFs model a distribution as flexible transformation of a base distribution, but are not applicable for VI due to dependence on already expressive base distributions. Our newly proposed \emph{Mixture of DNFs} (MDNF), however, can express any categorical distribution arbitrarily accurately (assuming sufficiently many mixture components) already with delta
base distributions (simplistic categorical distributions with all probability mass on one value). MDNF can be trained by gradient optimization, provide discrete samples following the true distribution, and -- in contrast to GS relaxation that fallaciously employs continuous probability density --
enables estimation of probability mass necessary for calculation of 
the variational objective. Compared to GS relaxation, MDNF is more reliable for variational approximation since it allows directly optimizing the true objective and does not have hyperparameters that are difficult to select. We demonstrate the first point in Figure~\ref{fig:bn_convergence}, and the latter later in Figure~\ref{fig:kl_mdnf_gumbel_annealing} (left).

We proceed by first discussing differentiable parameterizations for categorical distributions in general, and then cover their use for variational approximations. Before going to details, we summarize the main contributions: We
    (a) propose MDNF, new reliable differentiable parameterization for categorical distributions;
    (b) provide practical learning algorithms for fitting MDNF as variational approximation, including one building on boosting VI \citep{BoostingRefining};
    (c) show that MDNF is robust to its hyperparameters and provides unbiased estimate of the variational objective (in contrast to GS); 
    (d) show that MDNF outperforms GS relaxation for inference of ordinary graphical models, and is more reliable also for VAEs
    {and (e) introduce partial flows -- an extension to discrete flows by \citet{tran2019discrete} exhibiting better theoretical properties and empirical performance.}

\section{GRADIENTS OF CATEGORICAL SAMPLES}
\label{sec:background}

Without loss of generality,
we focus here on a one-dimensional categorical distribution, but in Section~\ref{sec:multivariate} explain how a set of $D$ distributions can be parameterized jointly while avoiding the naive storage cost of $K^D$.
A single categorical distribution is natively parameterized by the probabilities $p_k$ of $K$ possible outcomes, with  $\sum_{k=1}^K p_k = 1$. 
For the direct parameterization gradients of expectations like \eqref{eq:elbo} are undefined, 
but
next we explain differentiable alternatives: GS relaxation and DNF (both having other limitations).

\subsection{Gumbel-Softmax Relaxation}
\label{sec:gumbel}

The Gumbel-Softmax (or Concrete) relaxation~\citep{jang2017categorical,maddison2017concrete} 
consists of three parts: (1) Gumbel augmentation; (2) approximating $\text{argmax}$ with $\text{softmax}$; (3) discretizing samples by pushing them through straight-through (ST) operation~\citep{bengio2013estimating}. The third operation is often omitted since using 'relaxed one-hot' encodings allows for smoother optimization, though at the cost of needing to adapt the model for continuous samples. 
We use \textit{GS} to refer to the latter approach, denoting the discretized one by \textit{ST-GS} or \textit{ST-Gumbel}.

The value for the $k$th category of a sample $x$ is
\[
x_k =  \frac{\exp((\log p_k+g_k)/\tau)}{\sum_{j=1}^K \exp((\log p_j+g_j)/\tau)}, 
\quad g_j \sim \text{Gumbel}(0,1).
\]
The distribution is typically specified via logits $\lambda_k = \log p_k$ for unconstrained optimization,
and the temperature hyperparameter $\tau$ controls `magnitude' of gradient so that larger $\tau$ typically makes optimization easier while introducing more bias and for $\tau \rightarrow 0$ the bias (but also the gradient) disappears. 

The (unbounded) density of the GS distribution
\begin{equation}
p(x) = \tau^{K-1} (K-1)! \prod_{k=1}^{K} \left( \frac{p_k x_k ^{-\tau-1}}{\sum_{j=1}^K p_j x_j^{-\tau}} \right)
\label{eq:gumbel_density}
\end{equation}
is specified for continuous samples $x$, so that for any valid discrete sample (including $ST(x)$) it is $0$. 
Hence discrete entropy -- and variational objective -- is not defined (however, see Section~\ref{sec:evaluation} for discussion on how it can be estimated),
and to use GS relaxation for VI one needs 
to either (a) relax the prior itself using \eqref{eq:gumbel_density} with some arbitrarily chosen temperature $\tau_p$ 
or (b) use alternative learning objectives that are not valid bounds for the evidence (but may work in practice). The first choice introduces a second hyperameter 
and both approaches disconnect the internal learning objective from the desired one (Figure~\ref{fig:bn_convergence}).
These properties make GS relaxations unreliable for VI.

\begin{figure}[t]
\begin{center}
\includegraphics[width=0.975\columnwidth]{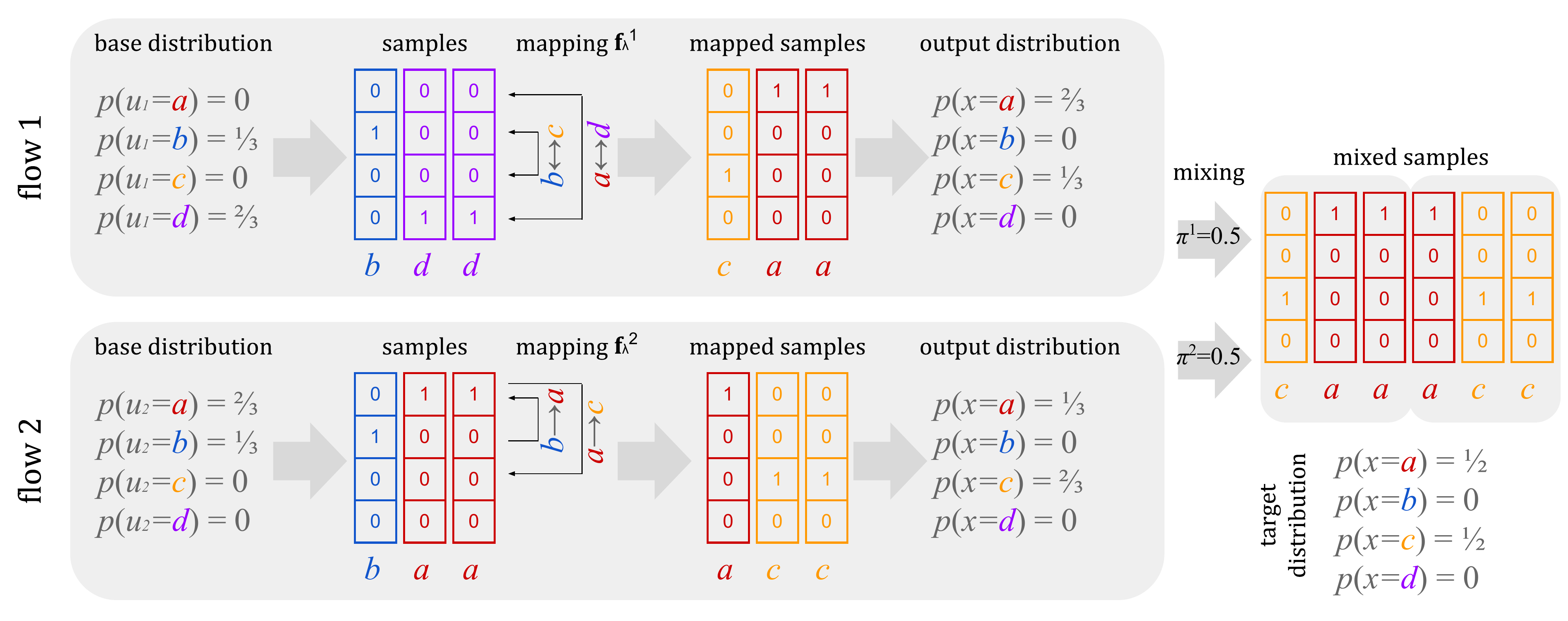}
\end{center}
\caption{A discrete normalizing flow constructed using \eqref{eq:transformation_tran} can only shuffle probabilities around, and is constrained by the choice of probabilities present in the base distribution, whereas
a mixture of flows with sufficiently many components can model any 
distribution also with simple base distributions.
}
\label{fig:mixture_of_flows}
\end{figure}

\subsection{Discrete Normalizing Flows}

An alternative parameterization for categorical distributions can be built using normalizing flows~(NF) \citep{tabak2010density,tabak2013family}. NFs transform  some (typically simple) base distribution $p_u(u)$ into the desired one via some invertible transformation with tractable Jacobian retaining the density.
With discrete normalizing flows~\citep{tran2019discrete}, the Jacobian is not needed and the probability is simply
$
p_x(x) = p_u(\invf(x))
$, but the choice of $\f(u)$ is considerably more limited -- we need an invertible discrete transtormation that can be trained w.r.t. the parameters $\lambda$.
\citet{tran2019discrete} proposed 
\begin{equation}
x := {\f}(u) = (\mu_{\lambda} + \sigma_{\lambda} \cdot u) \text{ mod } K,
\label{eq:transformation_tran}
\end{equation}
where $K$ denotes a number of possible categories 
and
$\sigma_{\lambda}$ and $K$ need to be coprime. 
The parameters 
$\mu_{\lambda}$ and $\sigma_{\lambda}$ are modeled with neural networks with $K$ outputs, scaled with a temperature $\tau$ and passed through softmax and the ST operation to obtain discrete one-hot encoded values.
That is, we have 
$\mu_{\lambda}= \text{ST}(\text{softmax}(\text{net}_\lambda(\dots)/\tau))$ for a suitable network.

\label{sec:limitations}
Unfortunately, {as also noted by \citet{papamakarios2019normalizing}}, the expressive power of DNF is seriously limited unless using already powerful base distributions. 
Since $\f$ is invertible and the samples $x$ and $u$ have the same shape, we have one-to-one mapping between $x$ and $u$. This implies that the possible values for target probabilities $p_x(x)$ are determined by the values $p_u(u)$ of the base distribution; for each unique $x$ with $p_x(x)=a$ for some $a$, we need $p_u(u)=a$ for some unique $u$. In other words, DNF can only move probability mass around without changing the values; see Figure~\ref{fig:mixture_of_flows} for an illustration. For specific constructions there can also be other limitations -- for example \eqref{eq:transformation_tran} cannot achieve all permutations -- but these can usually be alleviated by stacking multiple flows. Stacking, however, does not help with the more fundamental limitation of inability to change the probability values.

For generative tasks \citet{tran2019discrete} 
overcame the problem by first training strong autoregressive or factorized models and showed consistent but somewhat small improvement for DNF fine-tuning the distribution.
Training such flexible base distributions is not, however, possible in VI. We are not aware of any attempts of using DNFs for this purpose,
even though continuous normalizing flows are frequently used as approximations \citep{kingma2016improved}.

\section{MIXTURE OF DNF}
\label{sec:mixture_of_flows}

We present a novel composite parameterization for categorical distributions that combines multiple component distributions using a mixture formulation
\begin{equation}
p(x) = \sum_{b=1}^B \rho^b p^b(x), \label{eq:mixture}
\end{equation}
where the mixing weights $\rho^b \geq 0$ s.t. $\sum_b \rho^b = 1$ delegate partial responsibility of the total probability mass to individual component distributions $p^b(x)$.
The set of distributions that can be expressed using \eqref{eq:mixture} naturally depends on $B$
and the choice of component distributions, with some obvious special cases: With $B=1$ but arbitrary $p^b(x)$ we retain the direct parameterization, whereas with $B=K$ and $p^b(x=b)=1$ we can express any distribution with $K$ delta distributions.

We use \eqref{eq:mixture} for representation of arbitrary distributions by using DNFs as flexible and trainable component distributions. 
The resulting 
\emph{mixture of discrete normalizing flows} (MDNF) for parameterization of a categorical distribution defines the probability
\begin{equation}
    p_x(x) = \sum_{b=1}^B \rho^b p^b_u(\invf^{b}(x))
    \label{eq:mixture_of_discrete_flows},
\end{equation}
where we call $\f^b$ \emph{component flows}. For $B=1$ we get standard DNF, whereas for $B>1$
both sampling and probability evaluation are algorithmically more involved but remain tractable, as described next.

\subsection{Operations}
\label{sec:operations}

\emph{Forward sampling} for MDNF consists of three stages: (1)~choosing a flow $b \sim \text{Categ}(\rho^1, \dots, \rho^B)$, (2)~sampling 
$u \sim p_u^b(u)$ from the base distribution of the $b$th flow (which may be the same for all flows), and (3)~passing the sample through the flow to obtain the final sample: $x := \f^b (u)$. 
We assume that only the last step is differentiable, and hence the particular sample $x$ is a function of $\lambda$ associated \emph{only} with the particular $\f^b$. 
Direct implementation of this scheme, however, 
requires a separate dynamic computation graph associated with every sample $x$.
For a more efficient implementation using only a single graph, we can mask the outputs, conceptually similar to how masking is used in transformers or masked autoencoders.
First, a whole batch consisting of $n$ samples $\{u\}$ is passed through all flows $\f^b$. Then, flow outputs are multiplied with $B$ masks $\{M^b\}$ responsible for choosing which flow each individual sample is assigned to ($1$ appears exactly for one of the $B$ positions). Finally, the masked outputs are summed up using
$x := \f(\{u\}) = \f^1(\{u\}) \cdot M^1 \oplus \dots \oplus \f^B(\{u\}) \cdot M^B$.

Backward \emph{evaluation of probability} follows the general formulation in \eqref{eq:mixture_of_discrete_flows}, i.e.,
probability of a one-hot encoded sample $x$ 
is obtained by passing the sample through \emph{all} $B$  flows in reverse mode to obtain $B$ one-hot encoded samples $\{u^b: u^b := \invf^b(x)\}$ for which probabilities $p^b_u(u)$ can be evaluated by using the base distribution associated with the $b$th flow.

\subsection{Design}
\label{sec:design}

MDNF is compatible with (a) any choice of flows $\f^b$ and (b) any base distributions $p_u^b(u)$. Next, we discuss practical choices for these.

Flows constructed with \eqref{eq:transformation_tran} are not able to model all possible permutations without stacking multiple transformations and even then we still lack guarantees. Therefore, in Supplement we introduce the alternative of \emph{partial flows} that allow achieving all permutations by stacking $O(K \log^2 (K))$ transformations and perform better than \eqref{eq:transformation_tran}. 
However, with the delta base distributions explained below, already one layer of \eqref{eq:transformation_tran} is sufficient, and hence we used it in our experiments.

For ordinary DNFs it is crucial to use expressive base distributions, as explained in Section~\ref{sec:limitations}, but for MDNF we can use simple base distributions.
In fact, already
\emph{delta} distributions
that allocate all probability for a single category $u^*$ (that is, $p_u(u)=1 \iff u=u^*$) are sufficient, offering both computational advantage and easy theoretical analysis. Already a single flow of the form \eqref{eq:transformation_tran} (with fixed $\sigma=1$) can move the probability to the desired location by shifting with suitable $\mu_{\lambda}$, and with sufficiently large $B$ this is enough for modeling arbitrary distributions. Some other base distributions could be more efficient by requiring smaller $B$, but
we will show in Section~\ref{sec:experiments} that delta distributions are also a good practical choice.

In more detail, already with uniform weights $\rho^b = 1/B$ there exist flows $\f^b$ such that the absolute error for any target distribution $p_t(x)$ and any outcome $x$ can be bounded by
$
\left |\sum_{b=1}^B \rho^b p^b_u(\invf^{b}(x))
 - p_t(x)\right | \leq 1/B,
$
and hence the approximation converges when $B \rightarrow \infty$. This follows directly from approximating each $p_t(x)$ with a subset of the $B$ flows, chosen independently so that the proportion of the flows best matches it (see Supplement for details).
Note that for single DNF ($B=1$) the bound is trivial, and if allowing for free $\rho$ then $B=K$ is enough for zero error.

\section{MDNF FOR VI}
\label{sec:variational}

We use MDNF for variational approximation $q_x(x) \approx p(x|\D)$ of a model with categorical latent variables $x$ and observed data $\D=\{y\}$. To specify the distribution (for given $B$), we need to fix the weights $\rho^b$ and the flows $\f^b$, controlled by the parameters $\lambda^b$ of the network outputting the parameters $\mu_{\lambda d}$ and $\sigma_{\lambda d}$.

\subsection{Learning Algorithms}
\label{sec:algorithms}

The forward sampling of MDNF is differentiable w.r.t $\lambda$, and hence we can optimize the Monte Carlo estimate of a variational objective using reparametrized gradients, e.g., using \eqref{eq:elbo}.
Below we explain two ways of how it can be done.
The first assumes fixed weights, whereas the latter trains individual components one by one in a boosting fashion.
Both methods can be generalized for models with also continuous variables, for example, by plugging in also continuous NFs. 

\paragraph{VI on Flows (VIF)} 
The mixture is highly expressive already for fixed $\rho^b$. By fixing them to $1/B$, we can jointly train $\{\lambda^b\}$ for all flows with ordinary gradient ascent.
One optimization step consists of sampling $x$ as explained in Section~\ref{sec:operations} and estimating gradients of the objective following \eqref{eq:elbo}, where the Monte Carlo estimate of entropy relies on \eqref{eq:mixture_of_discrete_flows}. This algorithm is easy to implement, but simultaneous training of multiple flows is poorly identified {- any two flows can be swapped without effect and the mixture has $B!$ equivalent solutions what may cause problems with direct gradient-based learning.}

\paragraph{Boosting VI on Flows (BVIF)}
To avoid simultaneous training of competing flows, we turn the attention to existing literature on variational boosting,
originally developed for iteratively increasing flexibility of approximations (in context of continuous variables). 
Variational boosting algorithms construct posterior approximation of the form
$
    q_x(x) = \SUMFLOWS \rho^b q_\lambda^b(x),
$
where we intentionally match notation with \eqref{eq:mixture_of_discrete_flows}, so that in our case $q_\lambda^b(x) = p_u^b(\invf(x))$.
We adapt the algorithm of
\citet{BoostingRefining}, which proceeds by iteratively training the $\rho^{b+1}$ and $\lambda^{b+1}$ by keeping flows and weights up to $b$ fixed, using the objective
\begin{align*}
\BOUND &= 
(1-\rho^{b+1})\E_{q^{(1\dots b)}} \left[ \log p(y,x) - \log q^{(1 \dots b+1)}(x) \right] \\
&+  \rho^{b+1} \E_{q^{ b+1}} \left[
\log p(y,x) - \log q^{(1 \dots b+1)}(x)
\right],
\end{align*}
where $q^{(1\dots b)}$ denotes a mixture composed from components $1 \dots b$. The optimization consists of $B$ steps, each using a number of gradient ascent iterations.
{Note, however, that the method can be further extended to automatically find $B$, similar to \citet{giaquinto2020gradient},
for example if combined with a suitable
decreasing prior (e.g., stick-breaking process) for the mixing weights $\rho$.}

\paragraph{Boosting Baseline (BVI)}
Boosting 
can, in principle, be used to  learn a flexible approximation also without training the individual flows. For completeness, we explain a baseline algorithm doing this, to illustrate the importance of training the individual flows. For fixed component distributions $q^b$, each of which is a delta distribution (other choices could be used as well), we merely train the weights $\rho^b$ iteratively as above. This naturally requires large $B$ to work well.

\subsection{Multivariate Distributions}
\label{sec:multivariate}

Until now we have focused on individual 1D categorical distributions, but 
all the derivations so far are as well valid for multivariate $x$.
Multivariate posteriors can be decomposed into 1D distributions as $q(x) = q(x_1)\dots q(x_d|x_1 \dots x_{d-1})$,
where each of the distributions is modeled as a DNF with parameters $\mu_{\lambda d}$ and $\sigma_{\lambda d}$
expressed via neural networks.
Those networks \emph{model dependencies} between the dimensions $d$, typically either using autoregressive or bipartite design. In autoregressive networks, the transformation of the $d$th dimension depends on outputs of the preceding dimensions $1\dots d-1$, i.e.,
$(\mu_{\lambda d}, \sigma_{\lambda d}) := \text{net}_\lambda(x_1, \dots, x_{d-1})$, implemented using a recurrent network such as LSTM~\citep{hochreiter1997long} or a masked autoencoder~\citep{germain2015made}. We use the latter. \old{in this work.}
Bipartite networks \citep{dinh2014nice} divide the dimensions $\{d\}$ into two disjoint sets so that the second set depends only on the outputs for the first set.
This simplifies computation (no iterating over $d$ is required) but does not allow for modeling arbitrary dependencies (although stacking multiple flows with different subsets helps).
Regardless if autoregressive or bipartite, the methods can be used on the level of individual flows or between multiple MDNFs, each being responsible for modeling a 1D distribution. We used the former approach. 

Whenever comparing against GS, 
to be fair,  
we carry out experiments with factorized approximations $q(x)=q(x_1) \dots q(x_{D})$.
Then, it is sufficient to treat the parameters $\lambda$ directly as outputs (logits) of the network, e.g., $\text{net}_\lambda(\emptyset)=\lambda$. 
Finally, let us also note that arbitrary design of the neural networks $\mu_{\lambda d}$ and $\sigma_{\lambda d}$ allows to extend their inputs, for example, by including also observed data $y$ as
$(\mu_{\lambda d}, \sigma_{\lambda d}) := \text{net}_\lambda(x_1, \dots, x_{d-1}, y)$ or simply $\text{net}_\lambda(y)$, 
to achieve \emph{amortized inference}, e.g., with VAEs.

\subsection{Bias and Variance}
\label{sec:bias_and_variance}

As illustrated in Figure~\ref{fig:bn_convergence}, evaluation and monitoring of convergence is difficult for GS relaxations, due to discrepancy between the internal learning objective and the true variational bound. For MDNF, however, the internal learning objective is unbiased estimate of the true objective, and hence can be used for monitoring optimization and convergence, as well as for comparing models. 
Since \eqref{eq:mixture_of_discrete_flows} defines the probability $q(x)$, 
similarly as for the joint probability term $\mathbb{E}_{q(x)} \log p(\D,x) \approx \frac{1}{S} \sum_{x \sim q(x)} \log p(\D,x)$,
we can directly use an unbiased MC estimator also for entropy $H(q) \approx \frac{1}{S} \sum_{x \sim q(x)} \log q(x)$. This holds for both factorized and non-factorized approximations. For the latter, $\invf(x)$ transforms $x \rightarrow u$, where $p_u(u)$ may be (and in our case is) a factorized distribution for which $\log p_u(u) = \sum_d \log p_u(u_d)$. 

The variance of the estimate depends on the number of samples $S$ in \eqref{eq:elbo}, but is in practice very low. In Supplement we empirically show that even for extreme cases ($S=1$, large $B$) the variance is well below 1\% of the mean value. We also note that for the VIF algorithm we can reach zero variance with $S=B$, by \emph{deterministically} allocating individual samples to component flows in order, 
instead of sampling $b \sim \text{Categ}(\rho)$. 

\subsection{Hyperparameters}

MDNF has two main hyperparameters, the number of flows $B$ and the temperature $\tau$. We will show in Section~\ref{sec:hyperparameter_experiment} that selecting $\tau$ is considerably easier than selecting the corresponding pair of $\tau$ and $\tau_p$ for GS relaxation (Section~\ref{sec:gumbel}). In practice, following \citet{jang2017categorical}, we recommend annealing with $\tau_t = 
\tau e^{-\gamma t} $ for the $t$th iteration, where e.g. $\gamma=0.01$ and $\tau=10$ seem to work well in most cases.

Accuracy typically improves roughly monotonically with $B$ until saturating. The variational objective is comparable across $B$, and hence we can simply select the best choice after trying a few alternatives; in our case $B=40$ was deemed sufficient. $B$ also increases computation time. The detailed complexity depends on type of used $\f$ and the dimensionality $D$. For example, for an autoregressive flow a forward pass takes $O(B \cdot D)$ and a backward evaluation $O(B)$ assuming cost of one dimensional transformation $O(1)$.

\section{RELATED WORK}

\paragraph{Discrete Latent Variables}
Besides the approaches considered here, inference for categorical latent variables can be carried out by specialized model-specific algorithms
(e.g., \citet{rolfe2016discrete,vahdat2018dvae} for discrete VAE)
or marginalization.
For model-independent scenarios an alternative to reparameterization is the REINFORCE algorithm, with recent focus in reducing its variance~\citep{mnih2014,mnih2016,tucker2017rebar}. Besides VI, methods based on relaxation have also been used to enable Hamiltonian dynamics for MCMC~\citep{zhang2012continuous,nishimura2020discontinuous}.
Finally, the quantization-based methods for VAEs by
\citet{NIPS2017_7210, razavi2019generating} can be viewed as kind of relaxation, but with modified objectives. Since they neglect the prior altogether, they do not generalize to general approximations beyond VAE.

\paragraph{Normalizing Flows}
Normalizing flows~\citep{
van2018sylvester,
rezende2015variational,kingma2016improved} 
used for learning flexible posterior approximations and generative distributions for continuous variables have been recently generalized for discrete ordinal~\citep{hoogeboom2019integer} and categorical~\citep{tran2019discrete} variables, but only for generative tasks. 
Our approach, building on the latter, extends the scope for VI. MDNF 
resembles also some constructs for continuous normalizing flows, briefly described here even though both the motivation and details differ notably. \citet{papamakarios2019normalizing} proposed a mixture of flows and use an expression similar to \eqref{eq:mixture_of_discrete_flows}, but they rely on flows with restricted and non-overlapping (continuous) support in the $u$-space, and hence the mixture actually corresponds to piece-wise application of $B$ separate flows. 
%
{Our approach is also related to the recent mixture formulation, developed independently and in parallel by \citet{giaquinto2020gradient}. Their goal, however, was  to improve normalizing flows for posterior inference of continuous models and they did not address discrete latent variables in any way.}
Similarly, the RAD architecture by \citet{dinh2019rad} partitions the domain and uses a mixture model with piecewise invertible maps to learn better continuous distributions. 
Finally, 
MDNF shares also certain ideas with Stochastic Normalizing Flows~\citep{hodgkinson2020stochastic} in how sampling consists of a stochastic step (in our case selection of a flow) and a deterministic step with normalizing flows, and with Continuously Indexed Flows~\citep{cornish2019relaxing}
in challenging the bijection assumption. 

\paragraph{Variational Inference}
Compared to the recent alternative of \mbox{(semi-)implicit} approximations that can also be applied for discrete variables \citep{yin2018semi,titsias2019unbiased}, MDNF has the advantage of fitting the standard reparameterization framework and that it is directly applicable also with alternative objectives that rely on access to $\log q(x)$, such as \citet{dieng2017variational,domke2018importance}.
Finally, advances in boosting VI~\citep{BoostingBlackBox,BoostingGuo,BoostingOptimization,
BoostingUniversal} could be incorporated to improve the BVIF algorithm now based on \citet{BoostingRefining}.

\section{EXPERIMENTS}
\label{sec:experiments}

We validate our claims with experiments on Bayesian networks and VAEs that (a) show MDNF is more reliable than GS relaxation, due to unbiased internal objective and easier hyperparameter selection, (b) compare different learning algorithms for MDNF, and (c) verify that delta base distributions are a suitable choice.
All details of the experiments ({including complete code and data used to produce the results and plots})
are provided in Supplement, along with additional experiments on hyperparameters selection, variance of the gradients, partial flows, and a third model (Gaussian mixture model).

\begin{figure*}[t]
\begin{center}
\includegraphics[scale=0.40]{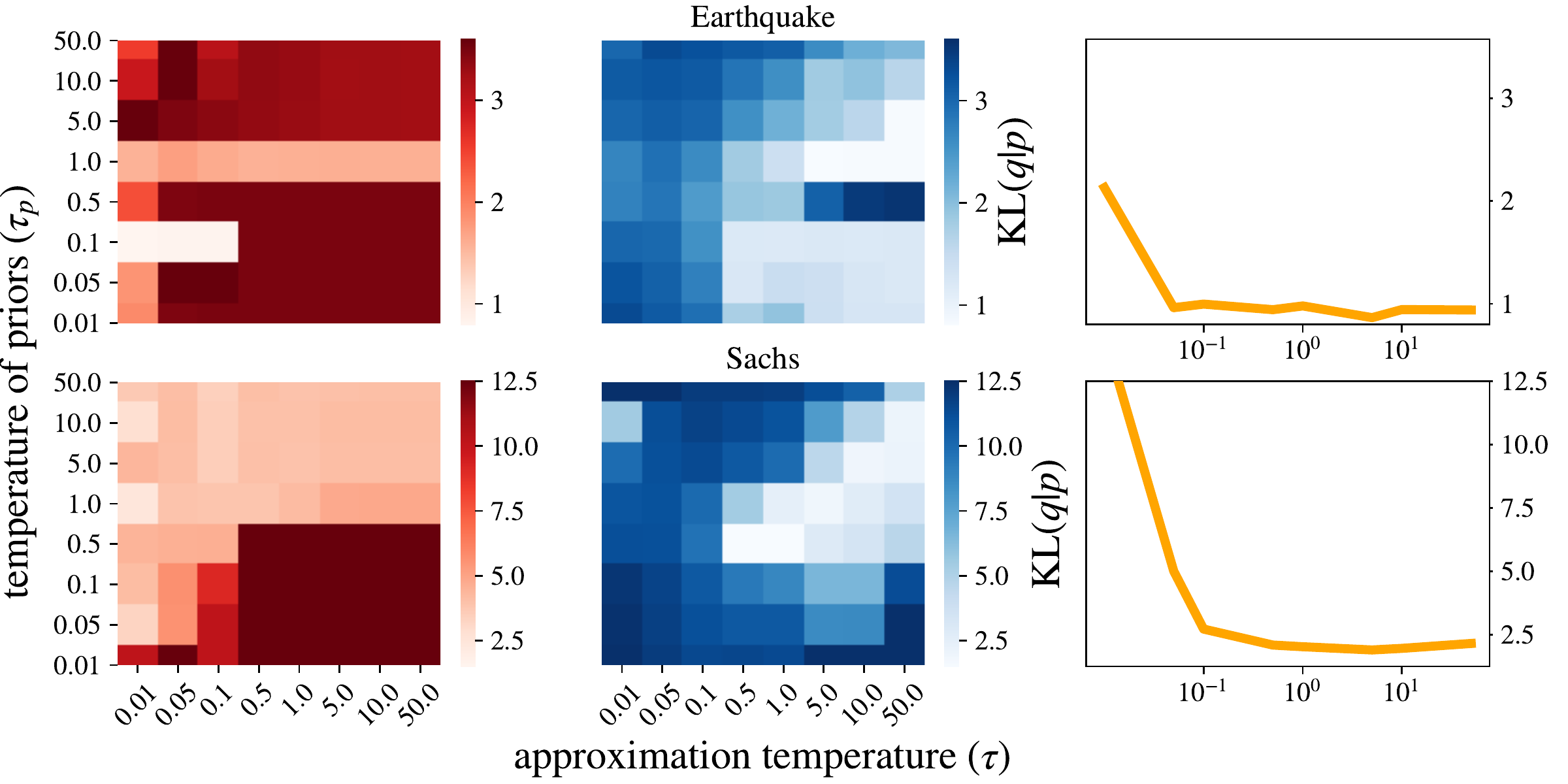}
\hspace{0.65cm}
\includegraphics[scale=0.40]{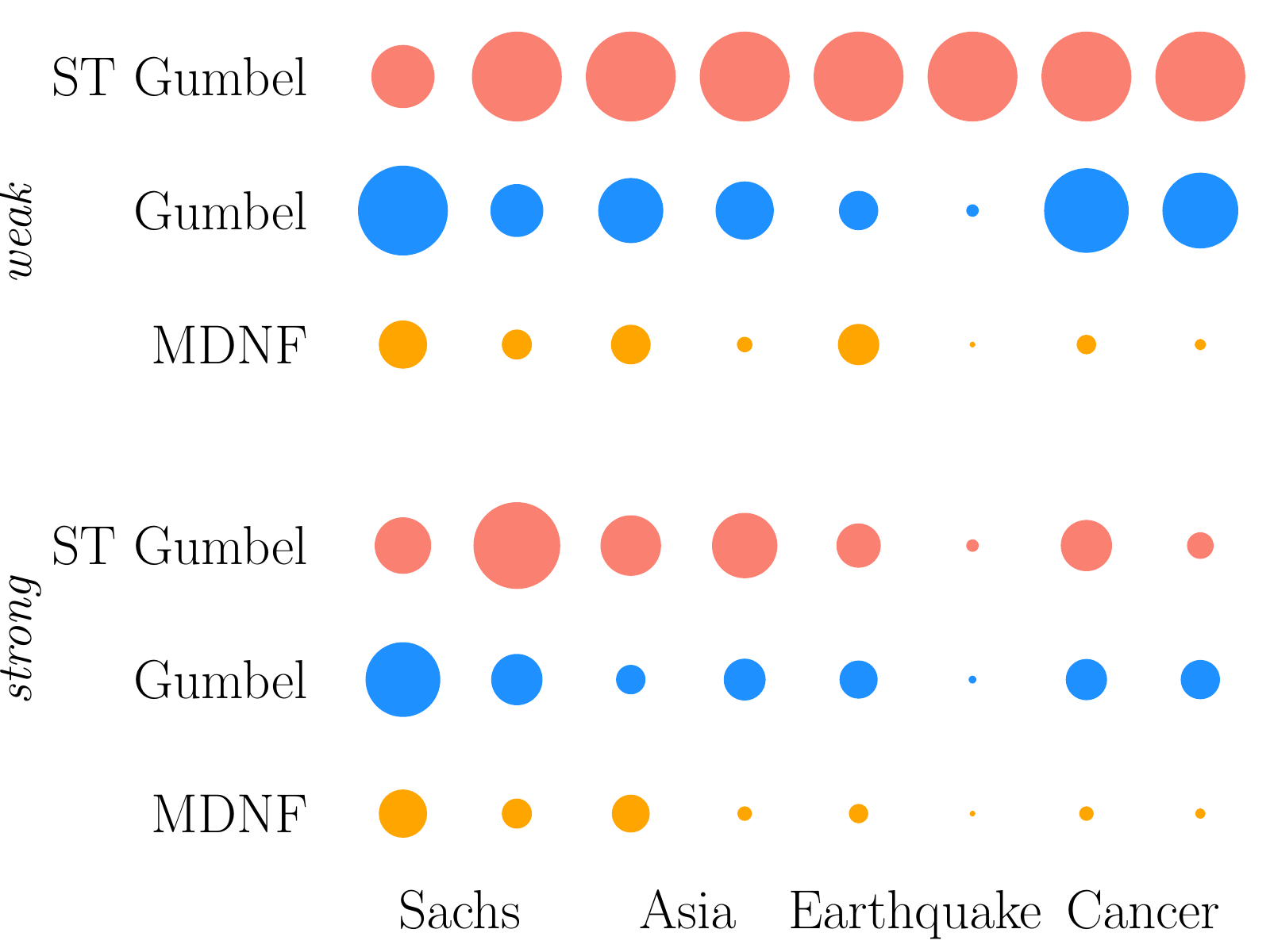}
\end{center}
\caption{
\textbf{(Left:)} 
KL divergences for a grid of
temperatures for two example BNs: \textit{Earthquake} (top) and \textit{Sachs} (bottom) for ST Gumbel Softmax (left; red), Gumbel Softmax (middle; blue) and MDNF (right; orange).  
\textbf{(Right:)} KL divergence (area;
the worst method has constant area for each column) 
with strong and weak oracles for selecting the temperatures, reported for four BNs and two choices of observed data for each. See Supplement for numerical values.
}
\label{fig:kl_mdnf_gumbel_annealing}
\end{figure*}

\subsection{Evaluation}
\label{sec:evaluation}

For evaluating the methods we use stochastic estimates for a variational learning objective, in practice ELBO, as well as the KL divergence to the true posterior that can be evaluated for sufficiently small models by exact inference. For MDNF the internal learning objective is unbiased estimate of ELBO, but for GS relaxation this is not the case. Instead, there can be a dramatic discrepancy between the two; Figure~\ref{fig:bn_convergence} compares the internal objective against the estimate described next.

We evaluate the ELBO for GS using empirical estimates obtained for large number of discretized ($\tilde{x}:=ST(x)$) samples -- this is an accurate, but computationally heavy, way of evaluating the approximation. The log-likelihood and log-prior terms of the objective can then be estimated using MC with $\frac{1}{S} \sum_{\tilde{x}} \log p(\D, \tilde{x})$, but the entropy requires estimating $\log q(\tilde{x})$. For factorized posteriors the entropy factorizes as $H(q) = \sum_{d=1}^D H(q_d)$, where each of the factors can be estimated as $H(q_d) = -\sum_{k=1}^K q_d(k) \log q_d(k)$, and for large $S$ asymptotically  $q_d(k) \approx \frac{\sum_{\tilde{x}} \textbf{1}[\tilde{x}=k]}{S}$.  \citet{archer2014bayesian} discuss alternative estimators for entropy.

\subsection{Reliability of Approximation}
\label{sec:hyperparameter_experiment}

\paragraph{Hyperparameters}
Using GS for variational approximation requires relaxing both the approximation and the prior \citep{maddison2017concrete} and hence GS has two tunable hyperparameters, $\tau$ and $\tau_p$, whereas MDNF uses discrete priors and only requires setting $\tau$. To illustrate the practical difficulty of using GS, we compare MDNF and GS relaxation in ideal conditions where an oracle evaluating the final quality against ground truth is available.

We apply the methods for posterior inference of discrete Bayes networks (BN) with multiple latent nodes (but fixed structure), so that for sufficiently small networks we can evaluate the true posterior by explicit enumeration (with exponential complexity). We use four networks from \url{https://www.bnlearn.com/bnrepository/}, so that always 1-2 variables are observed and others are latent, presenting the results for two choices for observed values for each BN. We use \textit{Asia} (8 binary nodes)
\textit{Sachs} (11 variables with 3 categories),
\textit{Earthquake} (5 binary nodes) and \textit{Cancer} (5 binary nodes). Optimization details for both MDNF and GS relaxation are provided in Supplement.

Figure~\ref{fig:kl_mdnf_gumbel_annealing} (left) illustrates the hyperparameter surfaces for two BNs.
For MDNF all temperatures above $\tau=1$ (we used here fixed temperatures, without annealing) have near identical performance, whereas for both GS variants there is a narrow range of  configurations that work well and the optimal choices are different for the two BNs. In summary, selecting good hyperparameters for GS is difficult even when having access to information that could never be available in practice. Note that for MDNF the choice can be based on the internal objective but for GS not -- the values across the rows would not be comparable due to different $\tau_p$.

Figure~\ref{fig:kl_mdnf_gumbel_annealing} (right) shows that even if one was able to select optimal hyperparameters by accessing an oracle, GS relaxation still does not work reliably in this task. The ST variant is uniformly bad, whereas the continuous variant is comparable with MDNF when using the optimal parameters selected specifically for this BN and task (\emph{strong oracle}). If allowed to access the true posterior but forced to use the same temperature parameters for all BNs and tasks (\emph{weak oracle}), it already starts performing poorly in many cases. For MDNF there is no notable difference, since  constant large $\tau$ is always near optimal.

\paragraph{Internal vs External Objective}
We already demonstrated in Figure~\ref{fig:bn_convergence} the difference between the internal objectives and the true quality, carried out with good temperature choices. For example, without ST on \textit{Sachs} network the variational objective starts growing when internal objective is still clearly improving. The illustrations here are selected examples observed during our experiments, and we note that in some cases the two objectives align a lot better. However, there is no way of identifying when this is the case, which means the method is in general fragile.

\subsection{Variational Autoencoder}
\label{sec:vae_experiment}

\begin{figure}[t]
\begin{center}
\includegraphics[width=0.425\textwidth]{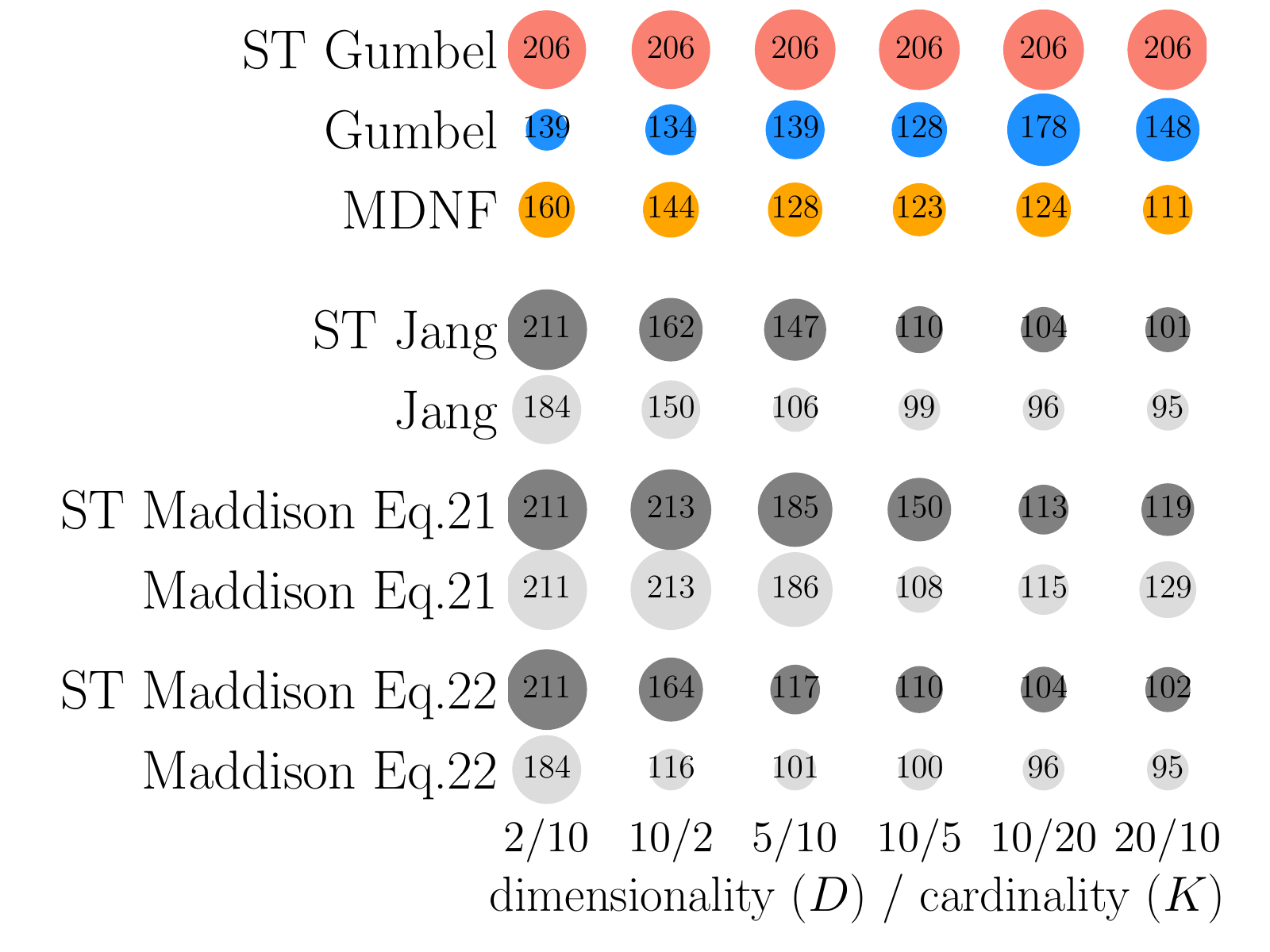}
\end{center}
\caption{Negative ELBO (lower is better; 
average over 3 random initializations) for different algorithms on VAEs of varying cardinality. The values are comparable between algorithms but not between models.
}
\label{fig:vae_elbos}
\end{figure}

\begin{figure*}[t]
\begin{center}
\includegraphics[width=0.24725\textwidth]{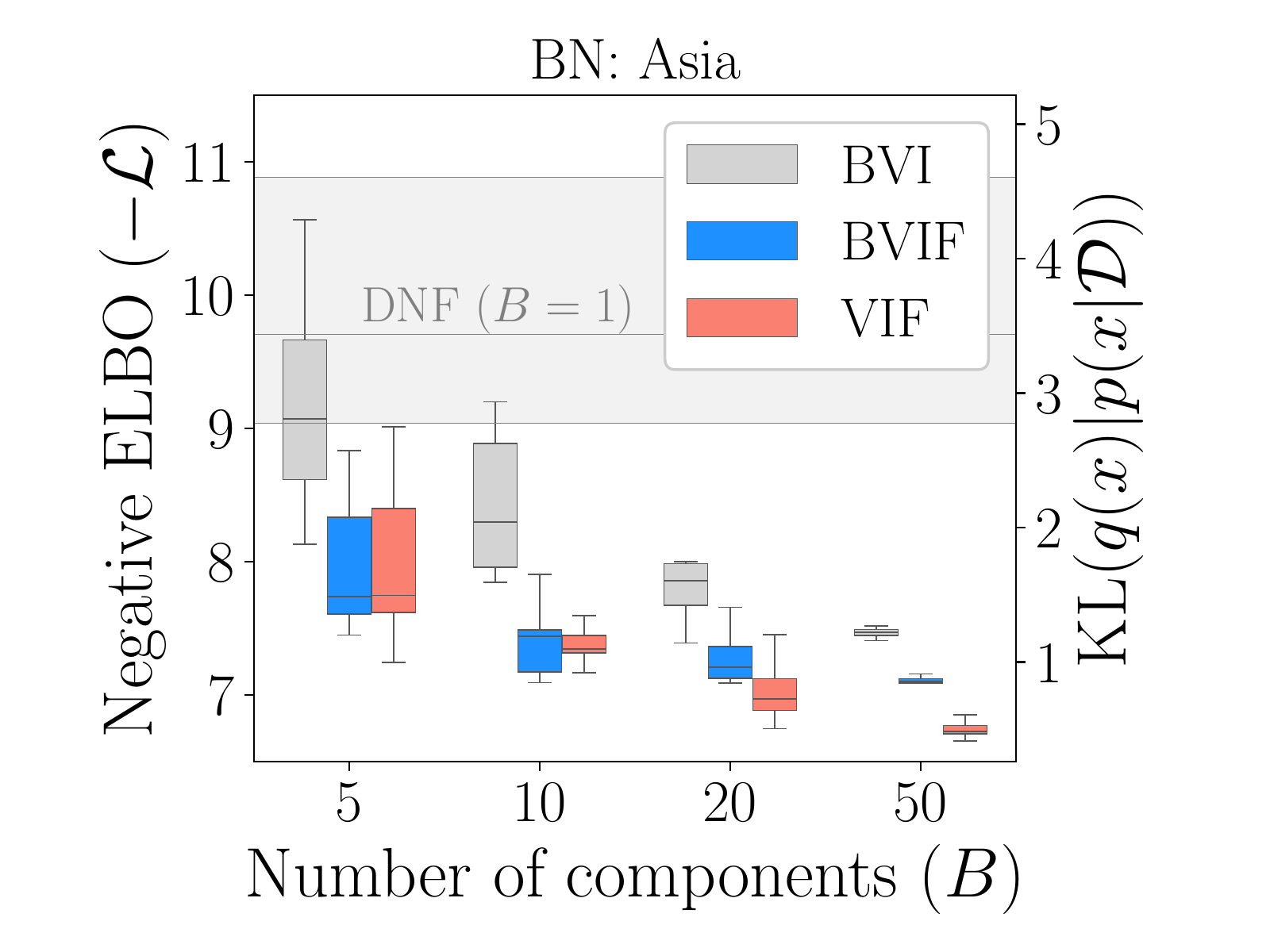}
\includegraphics[width=0.24725\textwidth]{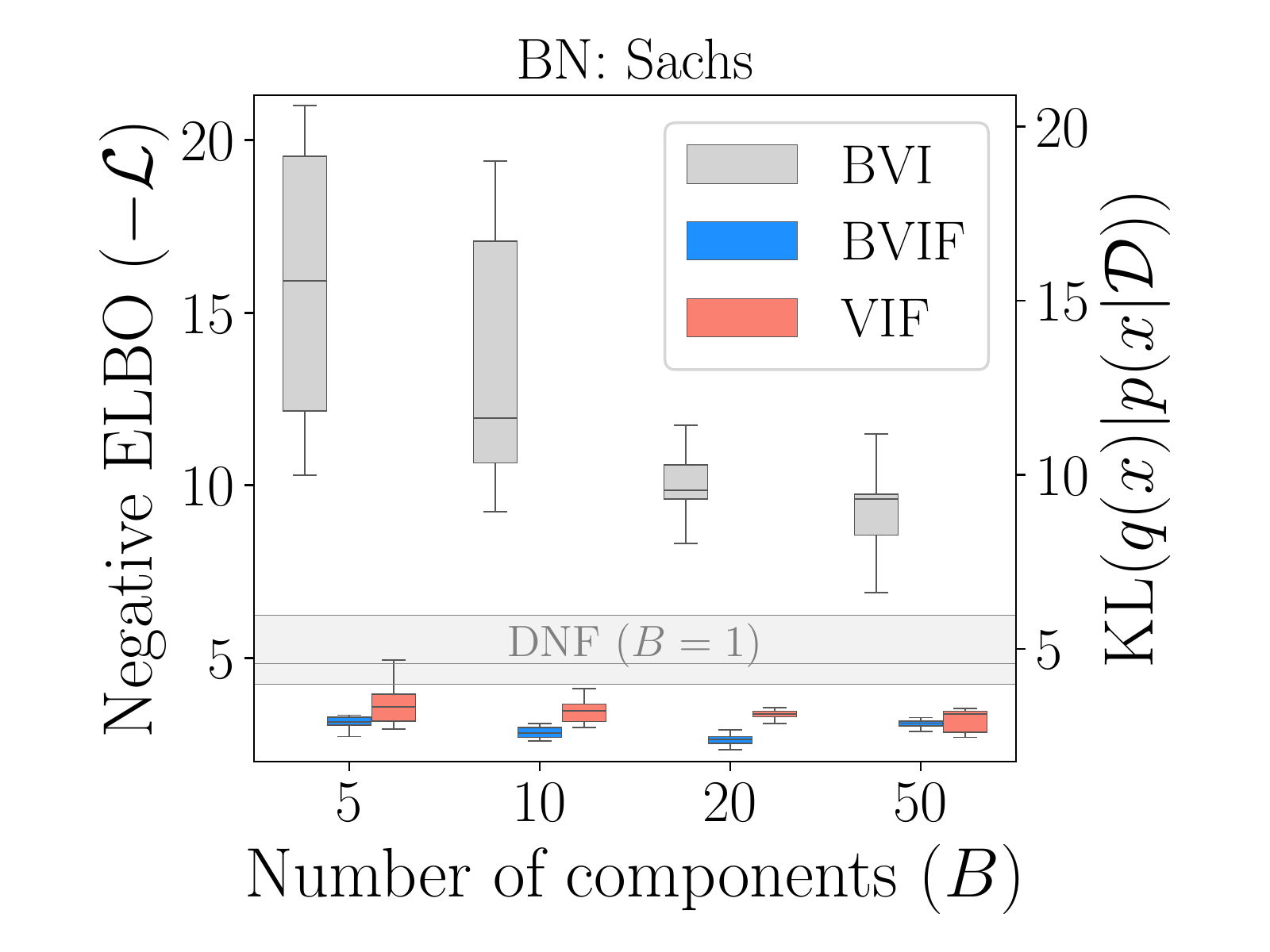}
\includegraphics[width=0.24725\textwidth]{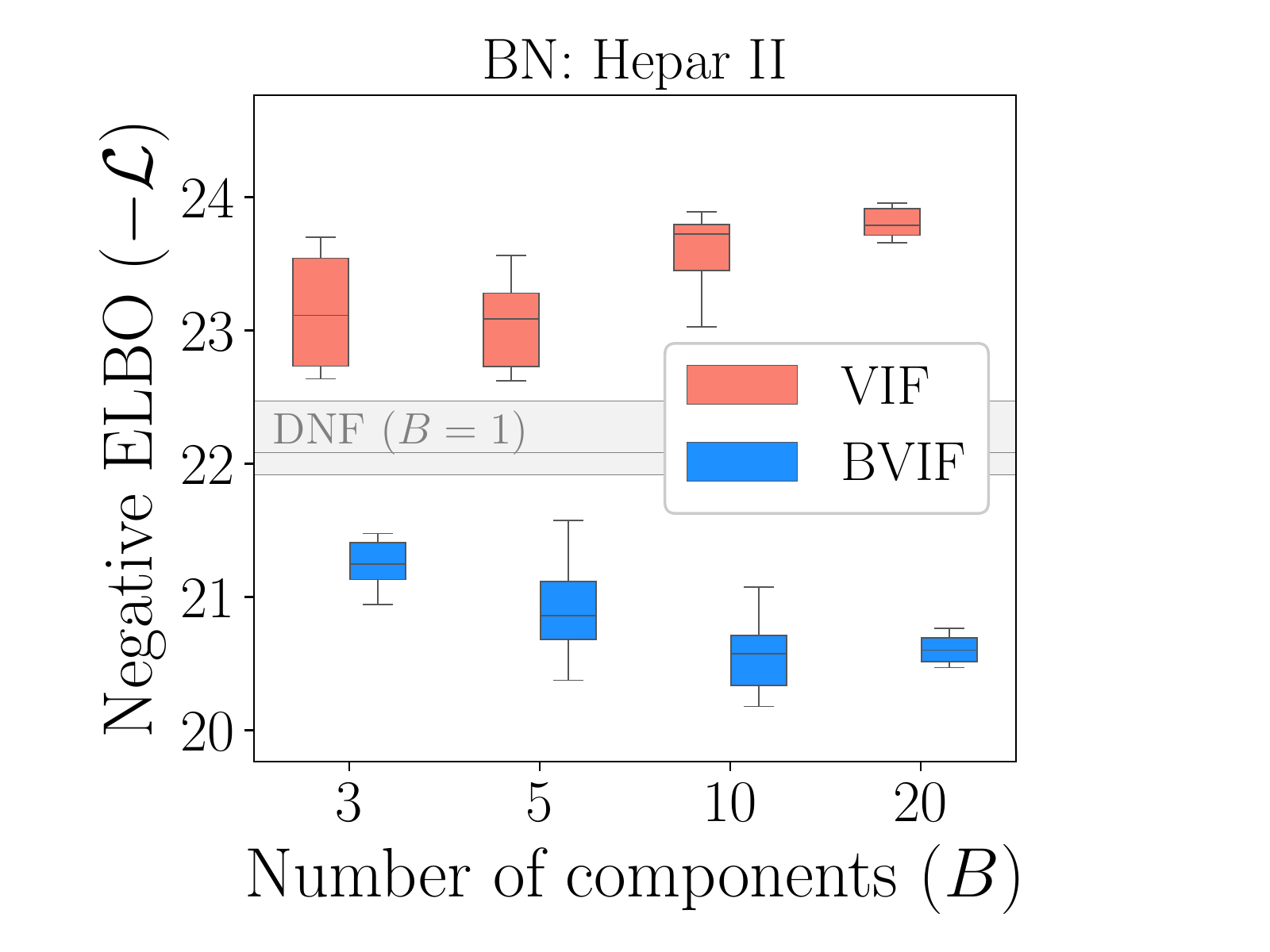}
\includegraphics[width=0.24\textwidth]{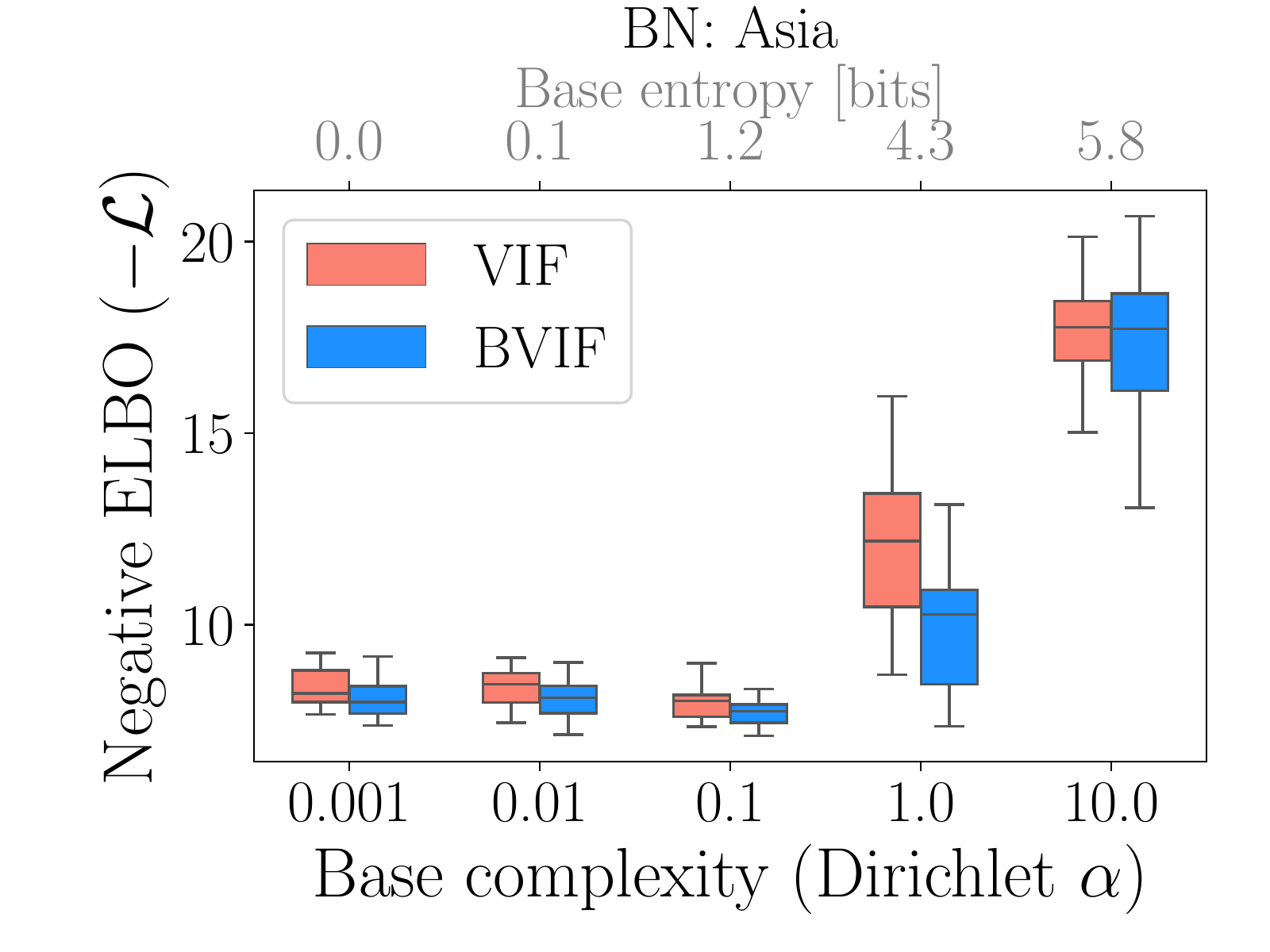}
\end{center}
\caption{({\bf Left three:}) Comparison of MDNF learning algorithms for three Bayesian networks.
{For \textit{Hepar II} BVI performs worse by an order of magnitude than the other algorithms and therefore, it was omitted from the plot.}
The box-plots indicate 25-50-75 percentiles over 10 repeated runs. The left axis show ELBO and the right axis KL divergence to the true posterior (zero is exact; not available for \textit{Hepar II}).
({\bf Right:}) Effect of base distribution on posterior accuracy; lower is better and small $\alpha$ corresponds to delta distribution. 
}
\label{fig:bayesian_networks}
\end{figure*}

Gumbel-Softmax relaxation is typically used for training Variational Autoencoders (VAEs) 
with discrete latent codes,
since the flexibility of the decoder can mask issues with the approximation and the choice of temperatures is less critical.
We compare MDNF ($B=40$, VIF algorithm, factorized approximation) against GS in this task.
{We omit the more computationally demanding BVIF from the experiment (even though it could perform superior to VIF) to avoid an unfair comparison where our method could spend more resources than the baselines.}

Besides \textit{Gumbel}, the only GS method optimizing a valid lower bound, 
we also compare against three heuristic alternatives often used for VAEs: \textit{Jang} refers to the objective of \citet{jang2017categorical} as implemented in \url{https://github.com/ericjang/gumbel-softmax}, and \textit{Madison Eq. X}, refer to two alternatives provided by the corresponding equations in \citet{maddison2017concrete}. For all methods we provide the results with and without ST, use hyperparameters suggested by the authors, and provide optimization details and visual illustrations 
in Supplement.

Figure~\ref{fig:vae_elbos} presents values of the sample-based variational objective (Section~\ref{sec:evaluation}) for all methods trained on MNIST digits for latent space configurations with increasing total cardinality (number of dimensions $D$ times the number of categories $K$). All methods except \textit{ST Gumbel} are in general competitive; they work well for at least some configurations. \textit{Gumbel} is the best for low total cardinality but performs poorly for high cardinality, whereas the opposite holds for \textit{Jang} and \textit{Maddison Eq. 22}. MDNF is the only method that works reliably and is close to the best method in all cases. That is, it is consistent and competitive in the tasks GS relaxation is most commonly used.

\subsection{Algorithms and Base Distributions}
\label{sec:algorithms_experiments}

Section~\ref{sec:algorithms} presented different algorithms for fitting MDNF. Figure~\ref{fig:bayesian_networks} (left three subplots) compares them on three BNs, including one larger example of \textit{Hepar II} (70 nodes with up to 6 categories).
BVIF outperforms the baseline of standard DNF ($B$=1) for all three BNs, and increasing $B$ improves the result roughly monotonically. For the two smaller networks VIF performs roughly as well, but breaks down for the larger \textit{Hepar II} network. BVI is clearly inferior, showing that training the component flows is necessary.

Throughout the paper we have used delta base distributions {allocating all probability mass for one outcome}. To study this choice in practice, we consider an alternative of sampling them from a symmetric Dirichlet $\text{Dir}(\alpha)$. For $\alpha \rightarrow 0$ all realizations are delta distributions and for $\alpha \rightarrow \infty$ they are uniform, and by varying $\alpha$ we can  create base distributions between these extremes. Note that $\alpha \rightarrow \infty$ is a worst-case baseline, since re-arranging identical values does not help. For both models very small $\alpha$ is optimal (Figure~\ref{fig:bayesian_networks} (right)), supporting use of delta distributions. 

Supplement replicates the above results for Gaussian mixture models (GMM) on three data sets to illustrate the method can also be used with continuous variables and with more latent variables. MDNF still outperforms DNF, but with marginal gain as the posterior is simpler: for GMMs all probability mass of a data point is often allocated to a single cluster.

\section{CONCLUSION}

Learning posterior approximations for high-dimensional categorical latent variables is fundamentally more difficult compared to continuous ones. Even though Gumbel Softmax relaxations are widely used for this in the context of VAEs, we showed that they are in practice fragile due to optimization objectives that can be severely biased compared to the true variational objective and because of high sensitivity to hyperparameters. For VAEs these problems are not very severe because using a flexible decoder makes them insensitive to the prior \citep{Zhao19}, but we demonstrated that even for VAEs 
our new alternative may be beneficial in terms of reliability (Figure~\ref{fig:vae_elbos}).
{MDNF consistently achieves competitive performance,  
while being more
reliable and easier to tune, offering a practical alternative for future applications.}

Our main interest was in posterior inference of proper probabilistic models, such as a discrete Bayes networks, for which Gumbel Softmax is poorly suited. We proposed an alternative reparameterization for categorical distributions -- MDNF, that (in practice) does not require fine-tuning of hyperparameters and provides better approximations.
{Unlike previous approaches, it enables direct probability evaluation and access to the true optimization objective.} \comment{We need to to remind about that again, since up to this point reviewers forgot about that}
MDNF builds on discrete normalizing flows, 
but can express arbitrary categorical distributions without pre-training the base distributions and hence  allows plug-and-play use for arbitrary models. 
We here used MDNF only for variational inference, but it 
can be used also 
for 
generative modeling tasks and has potential for improving on DNF.
{More generally, even though our work does not directly consider any application domain or task, it helps developing reliable models with discrete structures, which may better match intuition and reality.}
\comment{needed to fill-in few lines}

\begin{acknowledgements} 
This work was supported by the Academy of Finland Flagship program: Finnish Center for Artificial Intelligence, FCAI, and by the Technology Industries Finland and Erkko Foundation: Interactive Artificial Intelligence for Driving R\&D.
\end{acknowledgements}

\bibliography{references}

\begin{thebibliography}{43}
\providecommand{\natexlab}[1]{#1}
\providecommand{\url}[1]{\texttt{#1}}
\expandafter\ifx\csname urlstyle\endcsname\relax
  \providecommand{\doi}[1]{doi: #1}\else
  \providecommand{\doi}{doi: \begingroup \urlstyle{rm}\Url}\fi

\bibitem[Archer et~al.(2014)Archer, Park, and Pillow]{archer2014bayesian}
Evan Archer, Il~Memming Park, and Jonathan~W Pillow.
\newblock Bayesian entropy estimation for countable discrete distributions.
\newblock \emph{The Journal of Machine Learning Research}, 15\penalty0
  (1):\penalty0 2833--2868, 2014.

\bibitem[Batcher(1968)]{batcher1968sorting}
Kenneth~E Batcher.
\newblock Sorting networks and their applications.
\newblock In \emph{Proceedings of the April 30--May 2, 1968, spring joint
  computer conference}, pages 307--314, 1968.

\bibitem[Bengio et~al.(2013)Bengio, L{\'e}onard, and
  Courville]{bengio2013estimating}
Yoshua Bengio, Nicholas L{\'e}onard, and Aaron Courville.
\newblock Estimating or propagating gradients through stochastic neurons for
  conditional computation.
\newblock \emph{arXiv:1308.3432}, 2013.

\bibitem[Bishop(2006)]{bishop2006pattern}
Christopher~M Bishop.
\newblock \emph{Pattern recognition and machine learning}.
\newblock Springer, 2006.

\bibitem[Campbell and Li(2019)]{BoostingUniversal}
Trevor Campbell and Xinglong Li.
\newblock Universal boosting variational inference.
\newblock In \emph{Advances in Neural Information Processing Systems 32:},
  pages 3479--3490, 2019.

\bibitem[Cornish et~al.(2019)Cornish, Caterini, Deligiannidis, and
  Doucet]{cornish2019relaxing}
Rob Cornish, Anthony~L. Caterini, George Deligiannidis, and Arnaud Doucet.
\newblock Relaxing bijectivity constraints with continuously indexed
  normalising flows.
\newblock \emph{arXiv:1909.13833}, 2019.

\bibitem[Dieng et~al.(2017)Dieng, Tran, Ranganath, Paisley, and
  Blei]{dieng2017variational}
Adji~Bousso Dieng, Dustin Tran, Rajesh Ranganath, John Paisley, and David Blei.
\newblock Variational inference via chi upper bound minimization.
\newblock In \emph{Advances in Neural Information Processing Systems}, pages
  2732--2741, 2017.

\bibitem[Dinh et~al.(2015)Dinh, Krueger, and Bengio]{dinh2014nice}
Laurent Dinh, David Krueger, and Yoshua Bengio.
\newblock {NICE}: Non-linear independent components estimation.
\newblock In \emph{3rd International Conference on Learning Representations,
  {ICLR} 2015, Workshop Track Proceedings}, 2015.

\bibitem[Dinh et~al.(2019)Dinh, Sohl-Dickstein, Pascanu, and
  Larochelle]{dinh2019rad}
Laurent Dinh, Jascha Sohl-Dickstein, Razvan Pascanu, and Hugo Larochelle.
\newblock A {RAD} approach to deep mixture models.
\newblock \emph{arXiv preprint arXiv:1903.07714}, 2019.

\bibitem[Domke and Sheldon(2018)]{domke2018importance}
Justin Domke and Daniel~R Sheldon.
\newblock Importance weighting and variational inference.
\newblock In \emph{Advances in neural information processing systems}, pages
  4470--4479, 2018.

\bibitem[Germain et~al.(2015)Germain, Gregor, Murray, and
  Larochelle]{germain2015made}
Mathieu Germain, Karol Gregor, Iain Murray, and Hugo Larochelle.
\newblock {MADE}: Masked autoencoder for distribution estimation.
\newblock In \emph{International Conference on Machine Learning}, pages
  881--889, 2015.

\bibitem[Giaquinto and Banerjee(2020)]{giaquinto2020gradient}
Robert Giaquinto and Arindam Banerjee.
\newblock Gradient boosted normalizing flows.
\newblock \emph{Advances in Neural Information Processing Systems}, 33, 2020.

\bibitem[Gumbel(1954)]{Gumbel54}
Emil~Julius Gumbel.
\newblock Statistical theory of extreme values and some practical applications:
  a series of lectures.
\newblock Technical Report Number 33, US Govt. Print. Office, 1954.

\bibitem[Guo et~al.(2016)Guo, Wang, Fan, Broderick, and Dunson]{BoostingGuo}
Fangjian Guo, Xiangyu Wang, Kai Fan, Tamara Broderick, and David~B. Dunson.
\newblock Boosting variational inference.
\newblock \emph{ArXiv:1611.05559}, 2016.

\bibitem[Hochreiter and Schmidhuber(1997)]{hochreiter1997long}
Sepp Hochreiter and J{\"u}rgen Schmidhuber.
\newblock Long short-term memory.
\newblock \emph{Neural computation}, 9\penalty0 (8):\penalty0 1735--1780, 1997.

\bibitem[Hodgkinson et~al.(2020)Hodgkinson, van~der Heide, Roosta, and
  Mahoney]{hodgkinson2020stochastic}
Liam Hodgkinson, Chris van~der Heide, Fred Roosta, and Michael~W Mahoney.
\newblock Stochastic normalizing flows.
\newblock \emph{arXiv preprint arXiv:2002.09547}, 2020.

\bibitem[Hoogeboom et~al.(2019)Hoogeboom, Peters, van~den Berg, and
  Welling]{hoogeboom2019integer}
Emiel Hoogeboom, Jorn Peters, Rianne van~den Berg, and Max Welling.
\newblock Integer discrete flows and lossless compression.
\newblock In \emph{Advances in Neural Information Processing Systems}, pages
  12134--12144, 2019.

\bibitem[Jang et~al.(2017)Jang, Gu, and Poole]{jang2017categorical}
Eric Jang, Shixiang Gu, and Ben Poole.
\newblock Categorical reparameterization with {G}umbel-{S}oftmax.
\newblock In \emph{International Conference on Learning Representations}, 2017.

\bibitem[Kingma et~al.(2016)Kingma, Salimans, Jozefowicz, Chen, Sutskever, and
  Welling]{kingma2016improved}
Durk~P Kingma, Tim Salimans, Rafal Jozefowicz, Xi~Chen, Ilya Sutskever, and Max
  Welling.
\newblock Improved variational inference with inverse autoregressive flow.
\newblock In \emph{Advances in neural information processing systems}, pages
  4743--4751, 2016.

\bibitem[Locatello et~al.(2018{\natexlab{a}})Locatello, Dresdner, Khanna,
  Valera, and R{\"{a}}tsch]{BoostingBlackBox}
Francesco Locatello, Gideon Dresdner, Rajiv Khanna, Isabel Valera, and Gunnar
  R{\"{a}}tsch.
\newblock Boosting black box variational inference.
\newblock In \emph{Advances in Neural Information Processing Systems 31}, pages
  3405--3415, 2018{\natexlab{a}}.

\bibitem[Locatello et~al.(2018{\natexlab{b}})Locatello, Khanna, Ghosh, and
  R{\"{a}}tsch]{BoostingOptimization}
Francesco Locatello, Rajiv Khanna, Joydeep Ghosh, and Gunnar R{\"{a}}tsch.
\newblock Boosting variational inference: an optimization perspective.
\newblock In \emph{International Conference on Artificial Intelligence and
  Statistics}, volume~84 of \emph{Proceedings of Machine Learning Research},
  pages 464--472. {PMLR}, 2018{\natexlab{b}}.

\bibitem[Maddison et~al.(2017)Maddison, Mnih, and Teh]{maddison2017concrete}
Chris~J. Maddison, Andriy Mnih, and Yee~Whye Teh.
\newblock The concrete distribution: {A} continuous relaxation of discrete
  random variables.
\newblock In \emph{International Conference on Learning Representations}, 2017.

\bibitem[Miller et~al.(2017)Miller, Foti, and Adams]{BoostingRefining}
Andrew~C Miller, Nicholas~J Foti, and Ryan~P Adams.
\newblock Variational boosting: {I}teratively refining posterior
  approximations.
\newblock In \emph{International Conference on Machine Learning}, pages
  2420--2429, 2017.

\bibitem[Mnih and Gregor(2014)]{mnih2014}
Andriy Mnih and Karol Gregor.
\newblock Neural variational inference and learning in belief networks.
\newblock In \emph{International Conference on Machine Learning}, page
  II–1791–II–1799, 2014.

\bibitem[Mnih and Rezende(2016)]{mnih2016}
Andriy Mnih and Danilo~J. Rezende.
\newblock Variational inference for {M}onte {C}arlo objectives.
\newblock In \emph{International Conference on Machine Learning}, page
  2188–2196, 2016.

\bibitem[Nishimura et~al.(2020)Nishimura, Dunson, and
  Lu]{nishimura2020discontinuous}
Akihiko Nishimura, David~B Dunson, and Jianfeng Lu.
\newblock Discontinuous {H}amiltonian {M}onte {C}arlo for discrete parameters
  and discontinuous likelihoods.
\newblock \emph{Biometrika}, 107\penalty0 (2):\penalty0 365--380, 2020.

\bibitem[Papamakarios et~al.(2019)Papamakarios, Nalisnick, Rezende, Mohamed,
  and Lakshminarayanan]{papamakarios2019normalizing}
George Papamakarios, Eric Nalisnick, Danilo~Jimenez Rezende, Shakir Mohamed,
  and Balaji Lakshminarayanan.
\newblock Normalizing flows for probabilistic modeling and inference.
\newblock \emph{arXiv:1912.02762}, 2019.

\bibitem[Polson et~al.(2013)Polson, Scott, and Windle]{Polson13}
Nicholas~G. Polson, James~G. Scott, and Jesse Windle.
\newblock Bayesian inference for logistic models using {P}\'{o}lya-{G}amma
  latent variables.
\newblock \emph{Journal of the American Statistical Association}, 108\penalty0
  (504):\penalty0 1339--1349, 2013.

\bibitem[Razavi et~al.(2019)Razavi, van~den Oord, and
  Vinyals]{razavi2019generating}
Ali Razavi, Aaron van~den Oord, and Oriol Vinyals.
\newblock Generating diverse high-fidelity images with {VQ-VAE-2}.
\newblock In \emph{Advances in Neural Information Processing Systems}, pages
  14837--14847, 2019.

\bibitem[Renjith et~al.(2018)Renjith, Sreekumar, and
  Jathavedan]{renjith2018evaluation}
Shini Renjith, A~Sreekumar, and M~Jathavedan.
\newblock Evaluation of partitioning clustering algorithms for processing
  social media data in tourism domain.
\newblock In \emph{2018 IEEE Recent Advances in Intelligent Computational
  Systems}, pages 127--131. IEEE, 2018.

\bibitem[Rezende and Mohamed(2015)]{rezende2015variational}
Danilo Rezende and Shakir Mohamed.
\newblock Variational inference with normalizing flows.
\newblock In \emph{International Conference on Machine Learning}, pages
  1530--1538, 2015.

\bibitem[Rolfe(2016)]{rolfe2016discrete}
Jason~Tyler Rolfe.
\newblock Discrete variational autoencoders.
\newblock In \emph{International Conference on Learning Representations}, 2016.

\bibitem[Tabak and Turner(2013)]{tabak2013family}
Esteban~G Tabak and Cristina~V Turner.
\newblock A family of nonparametric density estimation algorithms.
\newblock \emph{Communications on Pure and Applied Mathematics}, 66\penalty0
  (2):\penalty0 145--164, 2013.

\bibitem[Tabak et~al.(2010)Tabak, Vanden-Eijnden, et~al.]{tabak2010density}
Esteban~G Tabak, Eric Vanden-Eijnden, et~al.
\newblock Density estimation by dual ascent of the log-likelihood.
\newblock \emph{Communications in Mathematical Sciences}, 8\penalty0
  (1):\penalty0 217--233, 2010.

\bibitem[Titsias and Ruiz(2019)]{titsias2019unbiased}
Michalis~K Titsias and Francisco Ruiz.
\newblock Unbiased implicit variational inference.
\newblock In \emph{The 22nd International Conference on Artificial Intelligence
  and Statistics}, pages 167--176, 2019.

\bibitem[Tran et~al.(2019)Tran, Vafa, Agrawal, Dinh, and
  Poole]{tran2019discrete}
Dustin Tran, Keyon Vafa, Kumar Agrawal, Laurent Dinh, and Ben Poole.
\newblock Discrete flows: {I}nvertible generative models of discrete data.
\newblock In \emph{Advances in Neural Information Processing Systems}, pages
  14692--14701, 2019.

\bibitem[Tucker et~al.(2017)Tucker, Mnih, Maddison, Lawson, and
  Sohl-Dickstein]{tucker2017rebar}
George Tucker, Andriy Mnih, Chris~J Maddison, John Lawson, and Jascha
  Sohl-Dickstein.
\newblock Rebar: Low-variance, unbiased gradient estimates for discrete latent
  variable models.
\newblock In \emph{Advances in Neural Information Processing Systems}, pages
  2627--2636, 2017.

\bibitem[Vahdat et~al.(2018)Vahdat, Macready, Bian, Khoshaman, and
  Andriyash]{vahdat2018dvae}
Arash Vahdat, William~G Macready, Zhengbing Bian, Amir Khoshaman, and Evgeny
  Andriyash.
\newblock {DVAE}++: Discrete variational autoencoders with overlapping
  transformations.
\newblock In \emph{International Conference on Machine Learning}, 2018.

\bibitem[Van Den~Berg et~al.(2018)Van Den~Berg, Hasenclever, Tomczak, and
  Welling]{van2018sylvester}
Rianne Van Den~Berg, Leonard Hasenclever, Jakub~M Tomczak, and Max Welling.
\newblock Sylvester normalizing flows for variational inference.
\newblock In \emph{34th Conference on Uncertainty in Artificial Intelligence},
  pages 393--402, 2018.

\bibitem[van~den Oord et~al.(2017)van~den Oord, Vinyals, and
  kavukcuoglu]{NIPS2017_7210}
Aaron van~den Oord, Oriol Vinyals, and koray kavukcuoglu.
\newblock Neural discrete representation learning.
\newblock In \emph{Advances in Neural Information Processing Systems 30}, pages
  6306--6315, 2017.

\bibitem[Yin and Zhou(2018)]{yin2018semi}
Mingzhang Yin and Mingyuan Zhou.
\newblock Semi-implicit variational inference.
\newblock In \emph{International Conference on Machine Learning}, pages
  5660--5669, 2018.

\bibitem[Zhang et~al.(2012)Zhang, Ghahramani, Storkey, and
  Sutton]{zhang2012continuous}
Yichuan Zhang, Zoubin Ghahramani, Amos~J Storkey, and Charles~A. Sutton.
\newblock Continuous relaxations for discrete {H}amiltonian {M}onte {C}arlo.
\newblock In \emph{Advances in Neural Information Processing Systems 25}, pages
  3194--3202, 2012.

\bibitem[Zhao et~al.(2019)Zhao, Song, and Ermon]{Zhao19}
Shengjia Zhao, Jiaming Song, and Stefano Ermon.
\newblock {InfoVAE}: Balancing learning and inference in variational
  autoencoders.
\newblock In \emph{AAAI Conference on Artificial Intelligence}, volume~33,
  pages 5885--5892, 2019.

\end{thebibliography}

\clearpage
\title{Reliable Categorical Variational Inference \\with Mixture of Discrete Normalizing Flows: Supplementary Materials}
\maketitle

In the Supplementary Materials, 
 we first provide additional illustrations and details of behavior of MDNF in Section~\ref{sec:DNF}, including demonstration of the newly proposed \emph{partial flows} that are useful also outside variational approximations. Then, in Section~\ref{sec:expdetails} we provide both the details of the experimental setup omitted from the main manuscript required for reproducing the experiments, and additional experimental results to complement the ones presented in the main paper.

\section{DISCRETE FLOWS AND MIXTURES}
\label{sec:DNF}

Here we illustrate in more detail how the DNF transformation proposed by \citet{tran2019discrete} works with delta base distributions (Section~\ref{sec:delta}), verify the approximation bound in Section~\ref{sec:design} 
(Section~\ref{sec:proof}), and explain and empirically validate the concept of \emph{partial flows} (Section~\ref{sec:comparison}).

\subsection{Delta Distributions and Shift Transformations}
\label{sec:delta}

\begin{figure*}[ht]
\begin{center}
\includegraphics[width=0.85\textwidth]{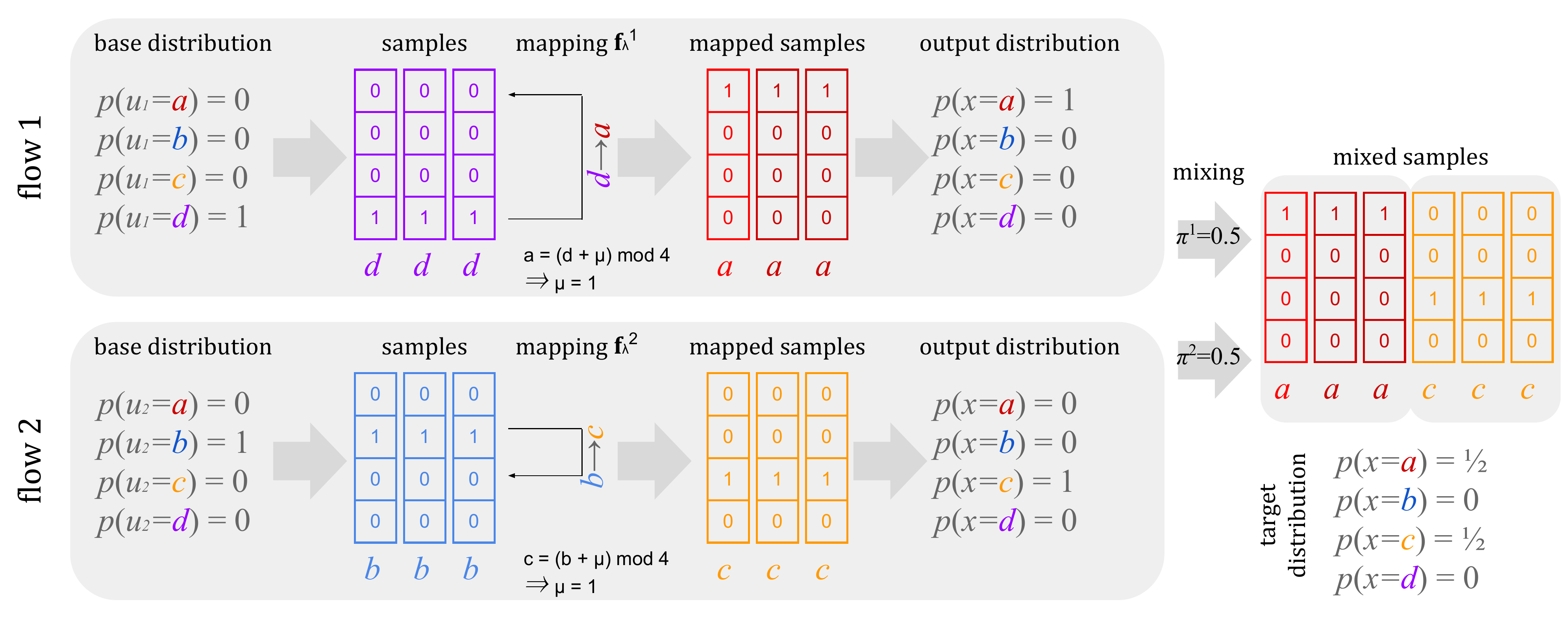}
\end{center}
\caption{Mixture of discrete normalizing flows with \emph{delta} base distributions and \emph{shift-only} transformations. }
\label{fig:delta_flows}
\end{figure*}

Even though MDNF can be used with any valid base distributions, 
delta distributions
have two concrete advantages: they do not require complex transformations and are easy to analyse. 
They allocate all probability mass for one entry, $p_u(c)=1$ for some $c$,
but
any delta distribution can be transformed into another one by renaming $c$ to some $k$ in samples drawn from the distribution. 
This can be achieved already with a single shift transformation -- 
by a simplified \eqref{eq:transformation_tran}
with $\sigma=1$ as
$$
x := {\f}(u) = (\mu_{\lambda} + u) \text{ mod } K.
$$
In particular, for any choice of $k$ and $c$, we can always find some $\mu$ such that $p_x(X=k)=1$ for any delta base distribution $p_u(U=c)$ using $k=(c+\mu) \text{ mod } K$.
In practice, we use trainable $\mu_\lambda = \text{ST}(softmax(\text{net}_\lambda(\dots)/\tau))$, and multiple flows in a mixture ($B>1$) to be able to allocate separately (but not  independently) fractions of the total probability mass.
Figure~\ref{fig:delta_flows} recreates Figure~\ref{fig:bn_convergence} but for delta base distributions with shift-only transformations.

\subsection{Error Bound for Delta Base Distributions}
\label{sec:proof}

As stated in Section~\ref{sec:design}, for any target distribution $p_t(x)$ over $K$ categories, there exist $B$ component flows $\f^b$ with mixture weights $\rho^b = 1/B$ such that
\[
\left |\sum_{b=1}^B \rho^b p^b_u(\invf^{b}(x))
 - p_t(x)\right | \leq 1/B
\]
for all $x$. As explained above, a single flow $\f^b$ using delta base distribution $p^b_u(\invf^b(x))$ can allocate all probability mass of that flow for any given category. To verify the bound, it hence remains to show that by combining $B$ component flows with uniform weights we can represent any categorical distribution sufficiently accurately. We do this in a constructive manner, and note that the reasoning does not directly say anything about how MDNFs trained with practical learning algorithms would behave. 

Let us first assume $K \le B$. We approximate $p_t(x)$ (for each $x$) using $P(x) \in \mathbb{N}_{0}$ flows, each carrying a probability mass of $1/B$. Let us first allocate $P(x) = \left \lfloor{p_t(x)B} \right \rfloor$ component flows for modeling each category, so that $p_t(x)$ is approximated by $\hat p_t(x) = \frac{P(x)}{B} = \frac{\left \lfloor{p_t(x)B} \right \rfloor}{B}$. For this it already holds that $|\hat p_t(x) - p_t(x)| \leq 1/B$. However, we have $B-K \le \sum_{x} P(x) \le B$, and hence $\hat p_t(x)$ is not a distribution and there are some unallocated component flows, unless the right inequality is equality (in which case $\hat p_t(x) = p_t(x)$ for all $x$). However, since the number of unallocated flows is at most $K$, we can allocate the remaining ones on arbitrary categories, one for each, so that $P(x) = \frac{\left \lceil{p_t(x)B} \right \rceil}{B}$ for these categories. This makes $\hat p_t(x)$ a valid distribution, while retaining the maximum approximation error of $1/B$.

If $K > B$ we simply set $P(x)=0$ for the $K-B$ smallest probabilities $p_t(x)$, all of which are guaranteed to be at most $1/K < 1/B$, and apply the above procedure to approximate $p_t(x)$ for the remaining categories.

\subsection{Partial Flows}
\label{sec:comparison}

\begin{figure}[t]
\begin{center}
\includegraphics[width=0.49\textwidth]{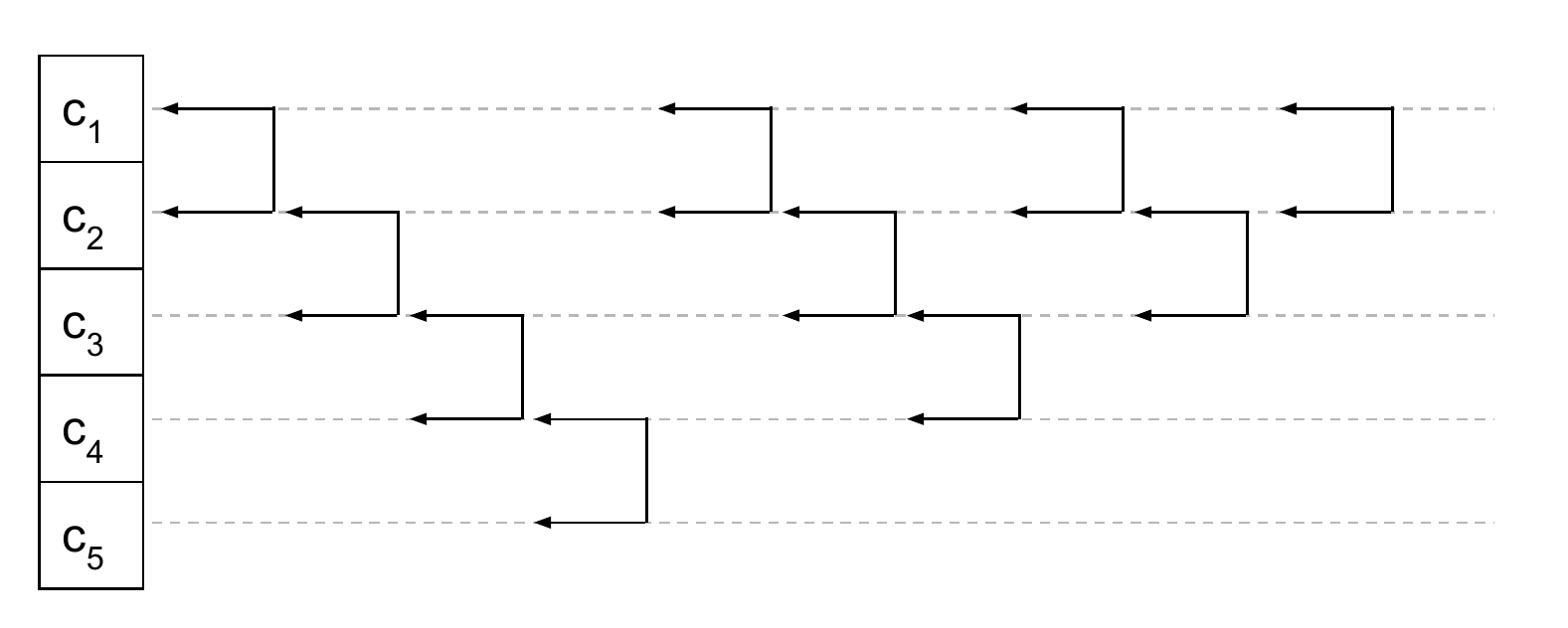}
\end{center}
\caption{Stack of 10 partial flows ($K'=2$) organized after bubble-sort sorting network for $K=5$ categories.}
\label{fig:bubble_sort}
\end{figure}

MDNF is in principle compatible with any choice of flows $\f^b$. In practice, the location-scale transformation of \citet{tran2019discrete} in \eqref{eq:transformation_tran} of the main manuscript works fairly well in approximation tasks and is used in the main paper, despite not being able to model all possible transformations without stacking multiple flows (and even then, we still lack guarantees).

For more general uses-cases, we propose an alternative of \emph{partial flows} that apply the transformation in \eqref{eq:transformation_tran} (usually with $\sigma \equiv 1$) but only on a subset of positions of a one-hot encoded vector $u_d$. Instead of acting on vector of length $K$ the transformation acts on vector of length $K'<K$. The remaining category positions are passed untouched. Note that the `subset' vectors do need to be valid one-hot encodings anymore and may consist of only $0$ in all positions.
 Such partial flows make, e.g., swapping elements considerably easier; in the extreme case of $K' = 2$ two elements can be swapped by shifting them by one position, without influencing any of the other probabilities (that would happen when doing the same swap with \eqref{eq:transformation_tran}).
 By simply stacking $O(K \log^2 (K))$ partial flows with appropriately chosen subsets of size $K'=2$, we can construct a flow that could perform any reordering (=relabeling) of categories by swapping individual pairs, analogous to bitonic sorter~\citep{batcher1968sorting}. 
Partial flows can be implemented with a slight modification of code for ordinary location-scale flows, e.g., by passing only the correct subset for an ordinary discrete flow transformation. Note however that transformation parameters $\mu$ need to receive full vectors $x_1, \dots, x_{d-1}$ when learning dependencies between dimensions.

\begin{table}[t]
  \begin{center}
    \caption{Comparison of stacks consisting of partial vs. location-scale flows in task of recovering the order of permuted categorical distribution with $K$ categories.}
    \label{tab:partial}
    \begin{tabular}{l|l|cc|cc} 
    \toprule
 &   & \multicolumn{2}{c}{$K=5$} & \multicolumn{2}{c}{$K=7$}  \\
 Flow & \#Layers	&	$p$	&	\#iters	&	$p$	&	\#iters\\
\midrule
\multirow{6}{*}{ \rotatebox[origin=c]{90}{Loc-scale} } & 3	&	0.38	&	268	&	0.08	&	1502\\
& 5	&	0.33	&	90	&	0.03	&	886\\
& 10	&	0.29	&	35	&	0.03	&	376\\
& 15	&	0.25	&	32	&	0.03	&	595\\
& 20	&	0.33	&	39	&	0.03	&	687\\
& 25	&	0.28	&	10	&	0.03	&	163\\
\midrule
\rotatebox[origin=c]{90}{Partial} & 10/21	&	1.00	&	78	&	0.78	&	512\\
\bottomrule
    \end{tabular}
  \end{center}
$p$ - fraction of successes  \\
\#iters - median of number of iterations
\end{table}

\paragraph{Experimental validation.}
To compare partial flows against location-scale flows of \eqref{eq:transformation_tran},  we run a controlled experiment with a 1-dimensional variable with $K=5/7$ categories 
distributed according to some $p_x$.
We used $p_x = [0.07, 0.13, 0.2 , 0.27, 0.33]$ with $K=5$ categories
and $p_x = [0.04, 0.07, 0.11, 0.14, 0.18, 0.21, 0.25]$ with $K=7$ categories. 
From the distribution $p_x$, we draw `features' (samples) $x$.
Likelihood of the samples $x$ is evaluated by
passing them in a reverse direction ($\invf$) through a stack consisting of a number of either partial of location-scale flows to get samples $u$ for which base probabilities $p_u$ are known.
For partial flows we always used $K'=2$ and organized them in a way resembling a bubble-sort sorting network (Figure~\ref{fig:bubble_sort}) so we used 10 flows for $K=5$ and 21 flows for $K=7$.
For location-scale flows we experimented with stacks of 3-25 identical layers.
In each of the repeated runs, we created a new base distribution $p_u$ by randomly shuffling $p_x$ and optimized the stacked flows to maximize the likelihood
by using Adam optimizer with learning rate 0.1 and $\tau=1.0$.
The optimal solution is reached only when categories of $p_u$ are correctly reordered back, so that samples $\f(u)$, $u \sim p_u$ follow $p_x$. This happens only for one of $K!$ permutations.

Table~\ref{tab:partial} presents results estimated with $40$ repeated runs, where as a success we count only the perfect recovery of the original probability distribution within at most $5000$ iterations. Partial flows are superior with both $K=5$ and $K=7$ categories, but require more iterations for convergence.
Additionally, for stacked location-scale flows, we observe that increasing the number of layers makes learning harder, i.e., the success is achieved in fewer cases (but also in fewer iterations).

\section{EXPERIMENTS}
\label{sec:expdetails}

In this section, we provide additional details of the main experiments presented in Section~\ref{sec:experiments}, including specification of models and networks used for transformations in MDNF. For most experiments we also present additional result figures and tables. Furthermore, we extend the experiment of Section~\ref{sec:algorithms_experiments} for another model family, namely mixture of Gaussians.

In all of the experiments,
we used delta base distributions, and therefore it was sufficient to use single-layer flows with shift-only transformation.
The shifts we found using standard gradient-based optimizers (RMSprop and Adam). %
In all cases,
following the common practice for Gumbel Softmax relaxations (see for example, \citet{jang2017categorical}), we also performed annealing of the temperature hyperparameter $\tau$ controlling bias of gradients of the straight-through estimator, by slowly decreasing its value in each iteration $t$ with $\tau_t = \tau \exp(-\gamma t)$, where $\gamma > 0$ controls the rate of annealing. An exception to this rule are the experiments on hyperparameter sensitivity, where we used constant temperatures to streamline the experiment and the presentation of the results. We replicated  those experiments with annealing but decided not to duplicate the result plots as the results are highly similar and all of the main conclusions hold also when using annealing.

\begin{table*}[t]
  \begin{center}
    \caption{
    Numerical values for Figure 3 (right):
 KL divergences with strong and weak oracles used for selecting the temperatures, reported for four BNs and two choices of observed data for each algorithm. Boldface indicates the best method for each case.
    }
    \label{tab:bn_numeric}
    \begin{tabular}{ll | p{1cm} p{1cm} p{1cm} p{1cm} p{1.85cm} p{1.85cm} p{1.65cm} p{1.65cm} } 
\toprule
 & & Sachs [akt=L] & Sachs [akt=H] & Asia [asia=y] & Asia [asia=y, xray=y] & Earthquake [marycalls=T] & Earthquake [marycalls=F] & Cancer [cancer=T] & Cancer [cancer=F] \\
\midrule
 \parbox[t]{2mm}{\multirow{3}{*}{\rotatebox[origin=c]{90}{\textit{weak}}}} & ST Gumbel & 3.70 & 9.30 & 6.51 & 9.75 & 2.55 & 6.07 & 2.64 & 0.68 \\
 & Gumbel & 2.76 & 8.18 & 0.59 & 4.30 & 0.83 & 0.07 & 1.38 & 1.03 \\
 & MDNF & 1.90 & \textbf{0.68} & 1.29 & 0.15 & 0.86 & \textbf{0.01} & 0.03 & 0.01 \\
\midrule
 \parbox[t]{2mm}{\multirow{3}{*}{\rotatebox[origin=c]{90}{\textit{strong}}}} & ST Gumbel & 2.49 & 8.11 & 6.51 & 4.15 & \textbf{0.80} & 0.05 & 0.87 & 0.11 \\
 & Gumbel & \textbf{0.97} & 1.70 & \textbf{0.55} & 2.41 & 0.81 & 0.02 & 0.35 & \textbf{0.00} \\
 & MDNF & 1.90 & \textbf{0.68} & 1.24 & \textbf{0.13} & 0.86 & \textbf{0.01} & \textbf{0.02} & 0.01 \\ 
\bottomrule
    \end{tabular}
  \end{center}
\end{table*}

\subsection{Reliability of Approximation (Complements Section~6.2)}

\paragraph{Details}
For this experiment
we used factorized posteriors with constant temperature for both MDNF and GS.
MDNFs were trained with VIF and the flows' shifts were represented directly as $\mu_d^b = \text{ST}(\text{softmax}(\lambda_d^b/\tau))$. 
Monte-carlo estimate (with $S=100$ samples) of ELBO 
we optimized
w.r.t. the parameters $\lambda^b$ of a MDNF
using RMSprop with learning rate $0.01$,
until convergence or for up to 10000 iterations.
In this experiment we used constant temperatures for all methods, simply to make the experiment easier to read. For completeness, we also repeated the experiment with temperature annealing. 
For MDNF annealing always helps, whereas for GS the behavior is inconsistent -- it may as well deteriorate results, but the differences are not significant.

\paragraph{Additional Results}
Table~\ref{tab:bn_numeric} presents the numerical results for the graphical summary presented in Figure~\ref{fig:kl_mdnf_gumbel_annealing} (right) of the main manuscript. For each of the four Bayesian networks, we fixed one or two of the variables to two alternative observed values (indicated in the table) and considered all other variables as unobserved.

\begin{figure*}[p]
\begin{center}
\includegraphics[scale=0.45]{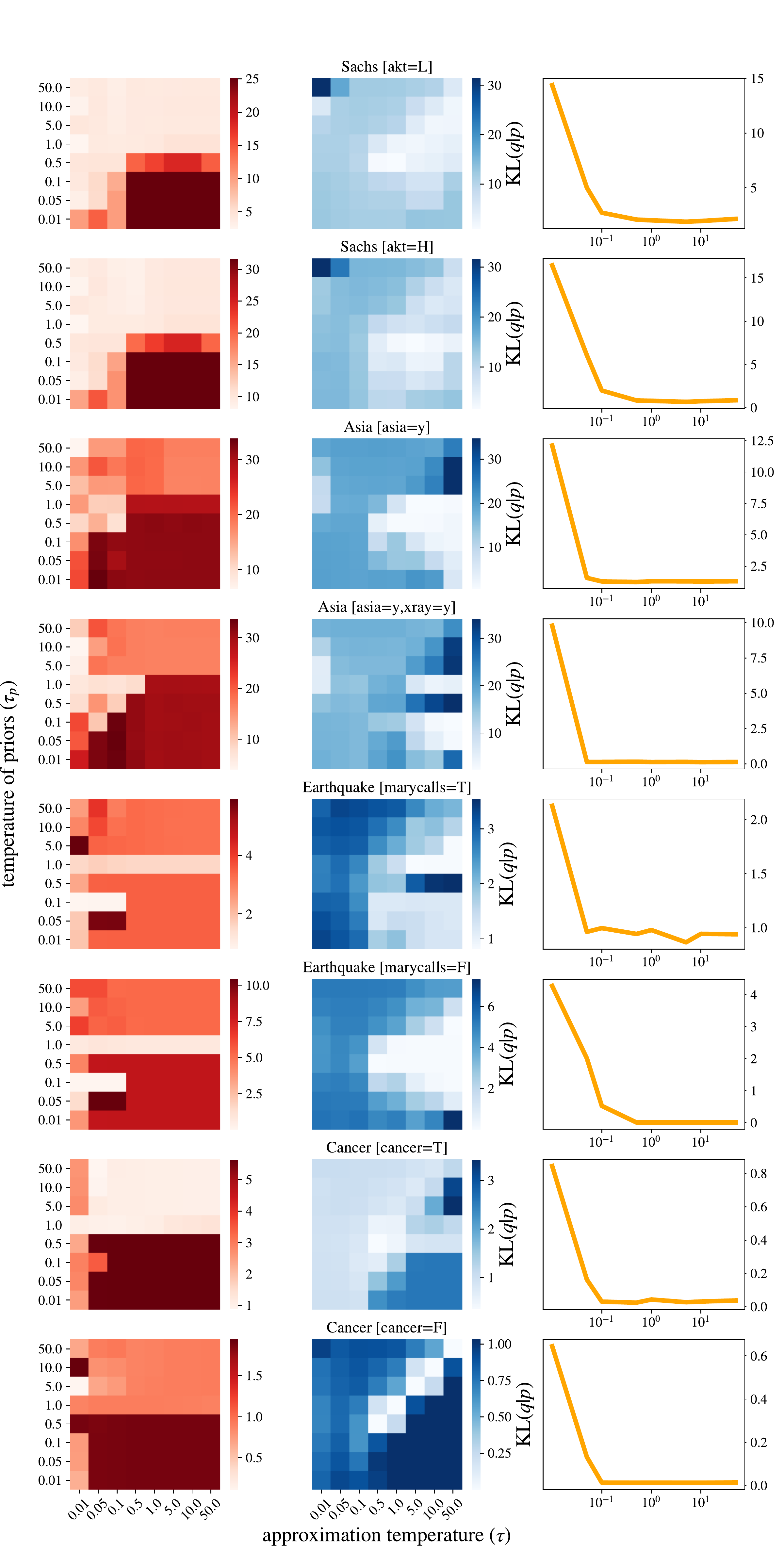}
\end{center}
\caption{
KL divergences for a grid of
temperatures for BNs for ST Gumbel Softmax (left; red), Gumbel Softmax (middle; blue) and MDNF (right; orange).  
}
\label{fig:kl_mdnf_gumbel_temperature_all}
\end{figure*}

Figure~\ref{fig:kl_mdnf_gumbel_temperature_all} extends the hyperparameter selection illustrations in Figure~\ref{fig:kl_mdnf_gumbel_annealing} (left), by providing the corresponding plots for all eight cases, replicating also the two cases provided in the main paper for ease of comparison. The main observations hold in all cases: Selecting the temperature for MDNF is always easy, whereas for Gumbel Softmax the surface is more complicated.

\subsection{Variational Autoencoders (Complements Section 6.3)}

\begin{figure}[ht]
\begin{center}
\includegraphics[width=0.5\textwidth]{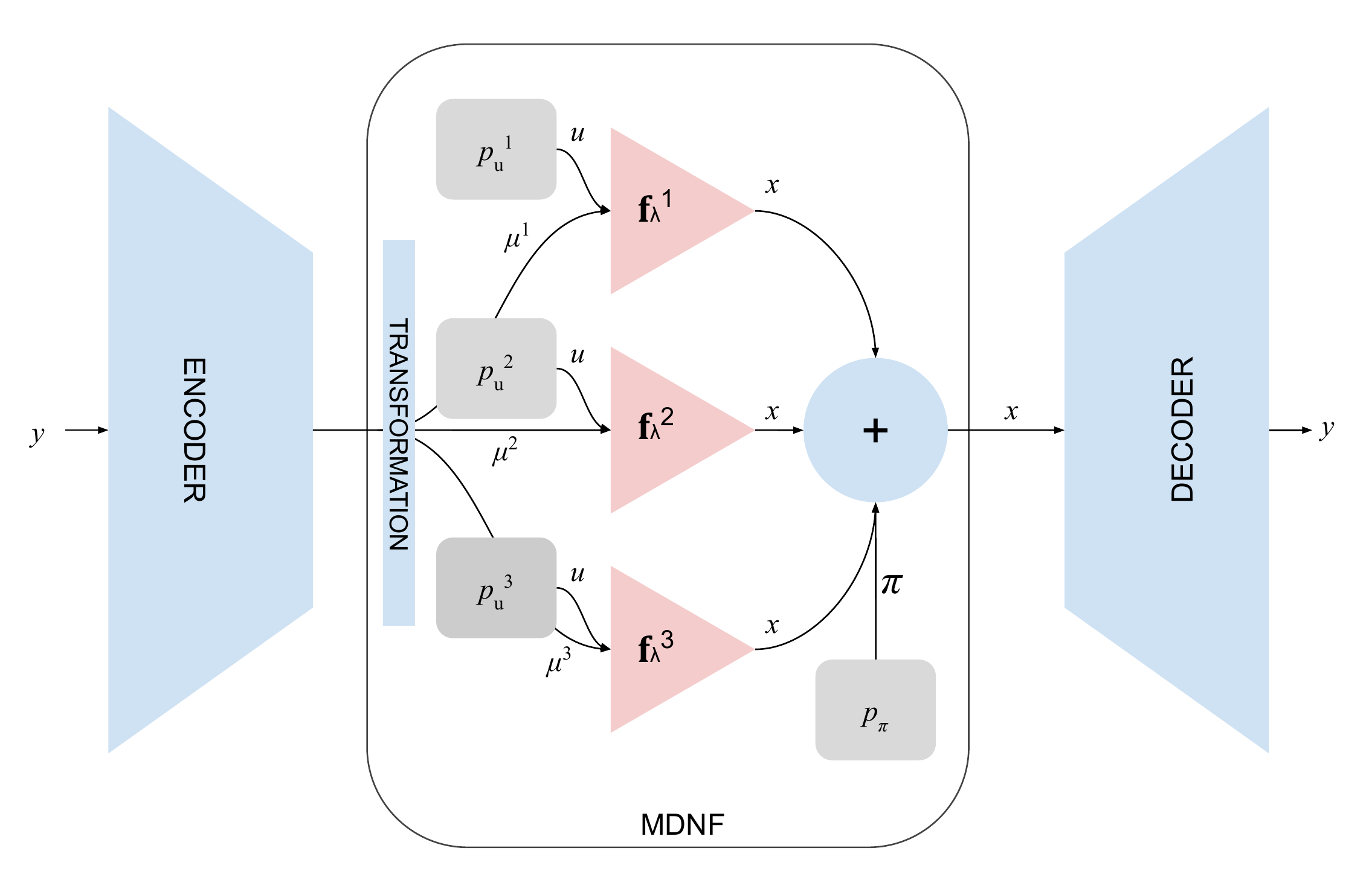}
\end{center}
\caption{MDNF-VAE: Variational autoencoder with mixture of  discrete normalizing flows (here $B=3$). }
\label{fig:mdnf_vae}
\end{figure}

\begin{figure}[ht]
\begin{center}
Gumbel \\
\includegraphics[width=0.29\textwidth]{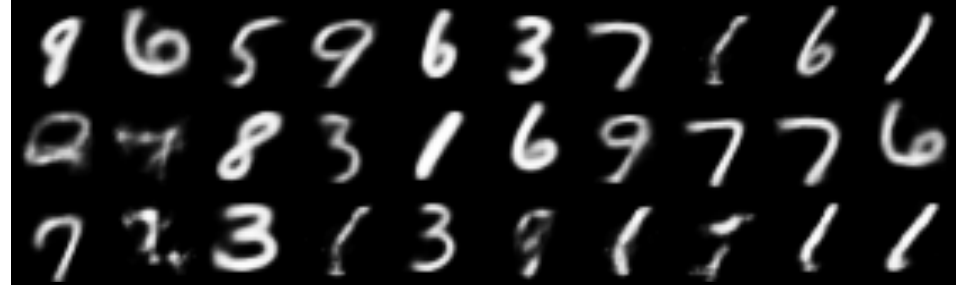}\\
ST Gumbel Softmax \\
\includegraphics[width=0.29\textwidth]{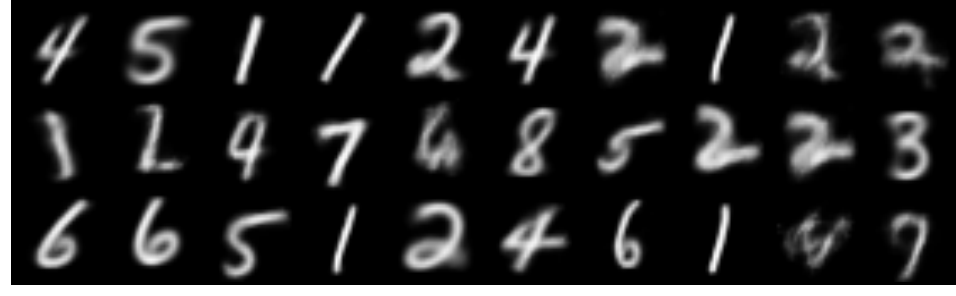}\\
MDNF (our)\\
\includegraphics[width=0.29\textwidth]{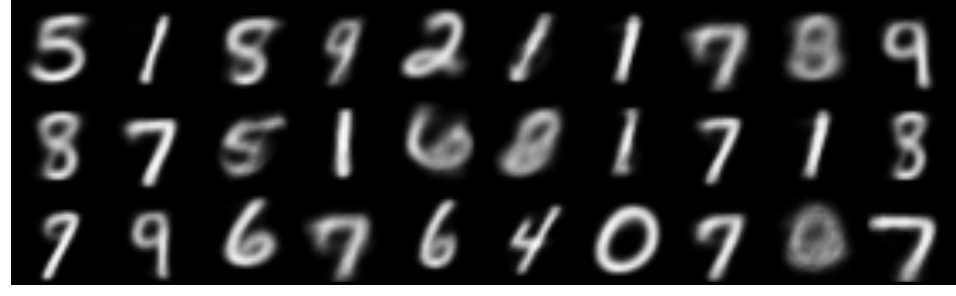}\\
\end{center}
\caption{Unconditional samples from VAEs with discrete latent variables ($D=10, K=2$) trained on MNIST.
}
\label{fig:vae}
\end{figure}

\paragraph{Details}

For the VAE experiment we used MDNF as illustrated in Figure~\ref{fig:mdnf_vae}.
The encoder's output $\lambda$ is passed to each of the component flows and 
parameterizes the flows' transformation $\mu^{b}(\lambda; \tau)$
(the transformations factorize over latent dimensions $d$). For the base distributions $p^b_u$ we used delta distributions, making it sufficient to use a single shift transformation $\mu^{b}$ with each of the $B=40$ used flows. 

The encoder and decoder architectures match the ones used by~\cite{jang2017categorical}\footnote{\url{https://github.com/ericjang/gumbel-softmax}}. The encoder has two dense layers with 512 and 256 nodes and ReLU activations, outputting $KD$ (for $D$ variables of $K$ categories each)
logits $\lambda$. The decoder has the same layers in reverse order, taking inputs of size $KD$ and outputting 768 logits for Bernoulli distributions for the pixels. The flow transformations  $\mu^{b}$ are obtained by passing the encoder's output through a network with a hidden layer consisting of $D \cdot K \cdot B$ nodes with ReLU activations and outputting tensors of the same shape. 

Our implementation of training follows closely the one by \citet{jang2017categorical}${}^1$: 
We used Adam with learning rate $0.001$,
minibatches of size $256$ and
uniform priors $p_x(x) = 1/K$, that for some of the Gumbel Softmax variants needed to be relaxed with $\tau_p=1.0$. For MDNF we used $\tau=100$, for GS we kept $\tau=1$, and the annealing was set to $\gamma=0.00003$.
Training of the VAE-MDNF we performed with VIF.

\paragraph{Additional Results}

To complement the numerical comparison in Figure~\ref{fig:vae_elbos}, we present examples of digits sampled from the models for one of the latent variable configurations in Figure~\ref{fig:vae}. We here illustrate the samples generated from MDNF and the primary competing method of Gumbel Softmax with relaxed priors; the samples for other variants are also very similar.

\subsection{Algorithms and Base Distributions (Complements Section 6.4)}

\begin{figure*}[t]
\begin{center}
\includegraphics[width=0.32\textwidth]{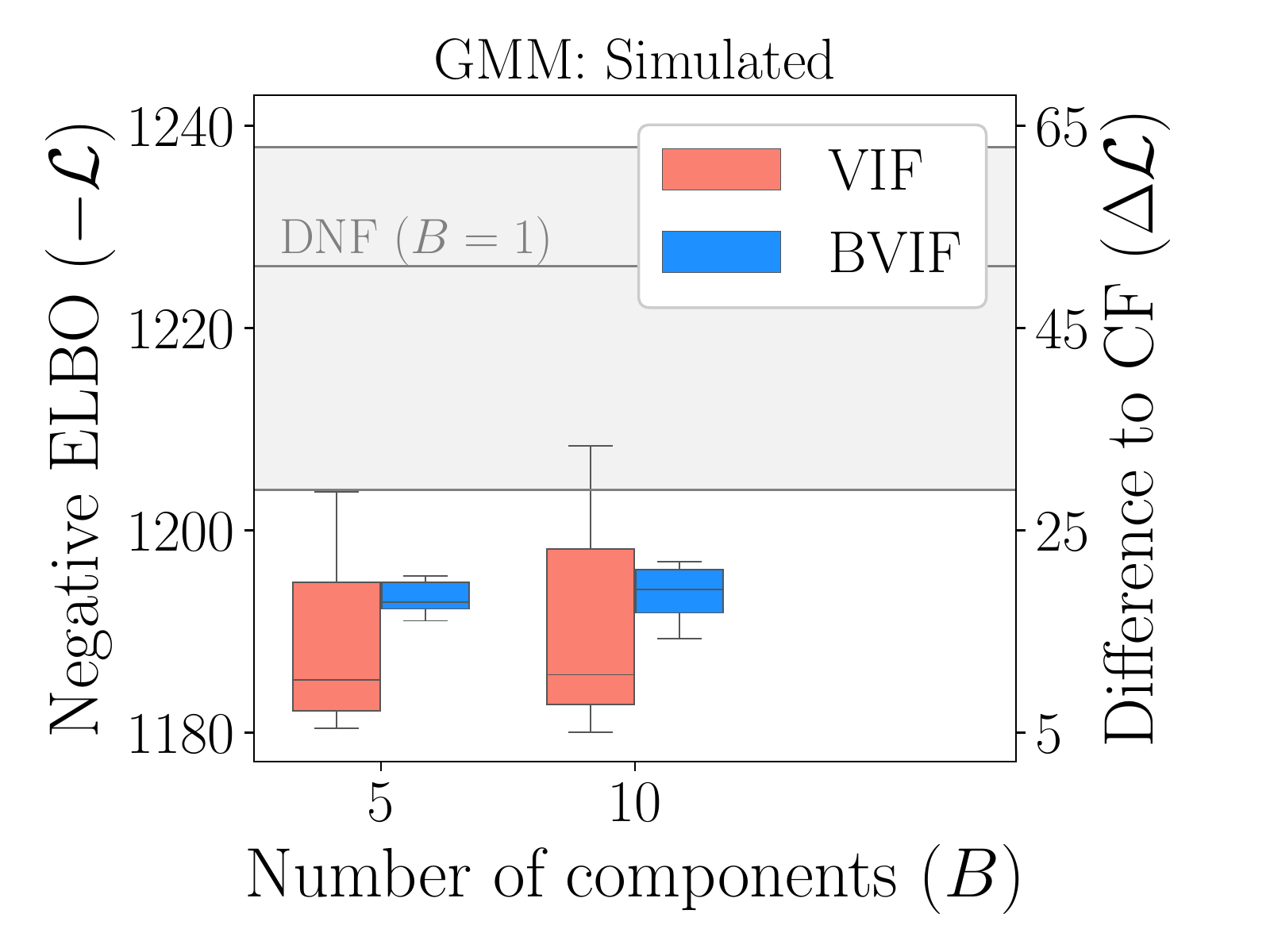}
\includegraphics[width=0.32\textwidth]{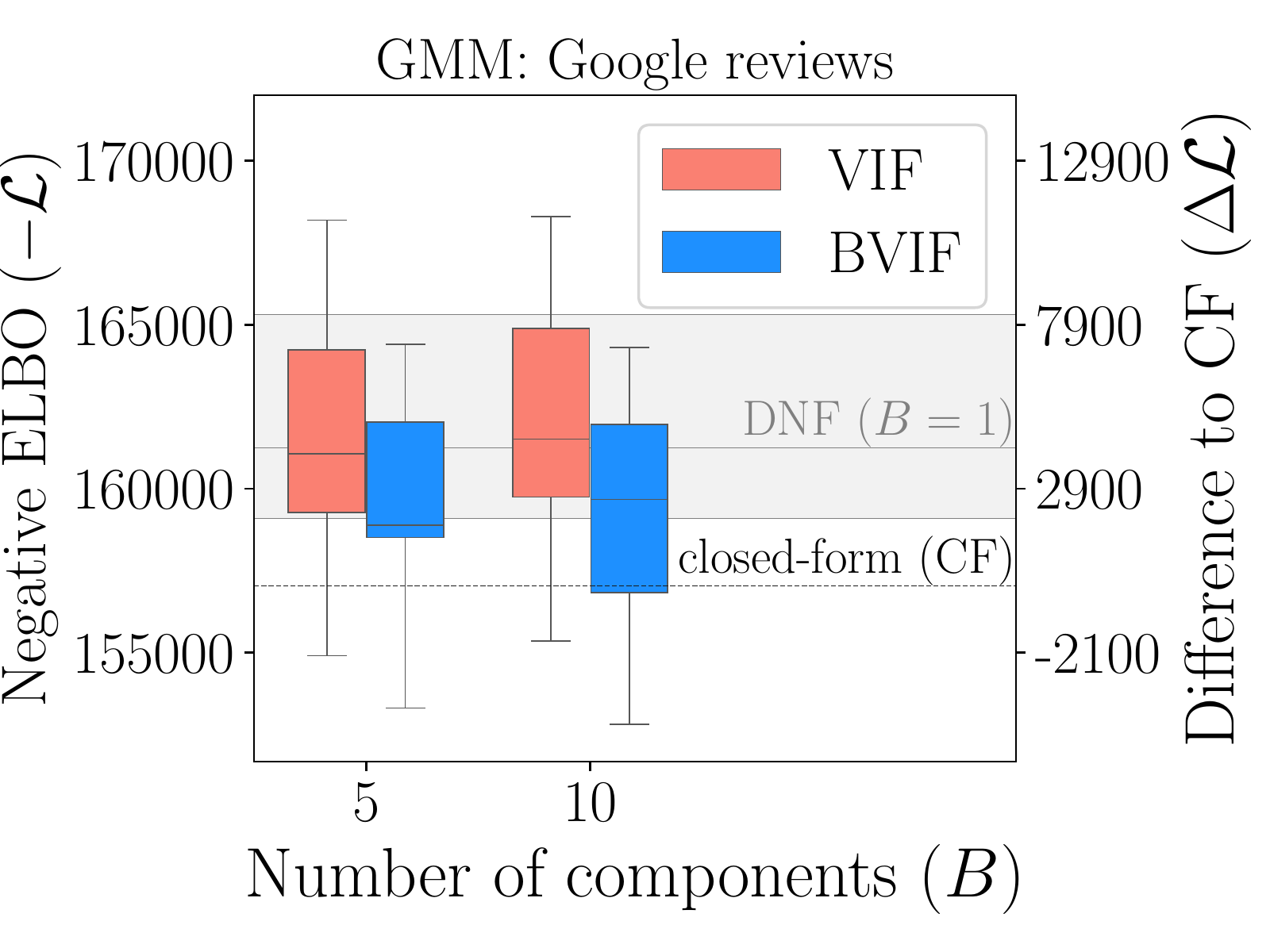}
\includegraphics[width=0.32\textwidth]{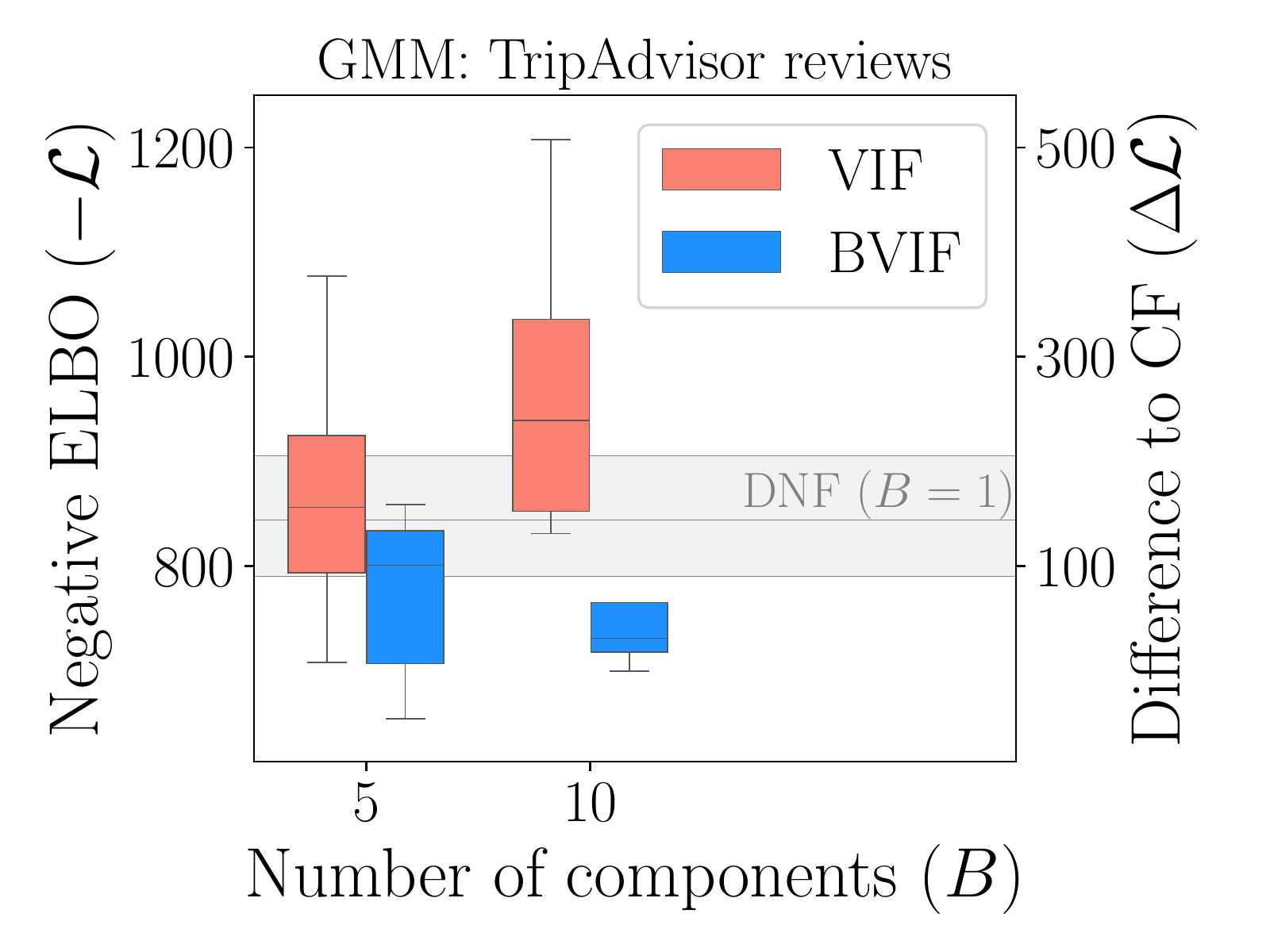}
\end{center}
\caption{Variational inference using MDNF for Gaussian mixture models on three data sets. 
The box-plots indicate 25-50-75 percentiles over 10 repeated runs. The right axis indicates difference to the closed-form (CF) solution.}
\label{fig:gmmvi}
\end{figure*}

\paragraph{Details}
In this experiment we used four small-to-medium sized publicly available\footnote{\url{https://www.bnlearn.com/bnrepository/}} networks for which the true posterior can be evaluated by direct enumeration (apart from \textit{Hepar}), observing values for 1-2 variables and leaving all others latent. We used \textit{Asia} (8 binary nodes; we fix \code{asia:=yes} and \code{xray:=yes}), \textit{Sachs} (11 variables with 3 categories; we fix \code{Akt:=LOW}) and \textit{Hepar II} (70 nodes with up to 6 categories; we fix \code{carcinoma:=present}).
To approximate the posteriors, 
we used masked autoencoders~\citep{germain2015made} (MADE) representing flows' shifts as $\mu^b_d = \text{ST}(\text{softmax}(\text{MADE}_{\lambda^b}(x_1, \dots, x_{d-1})/\tau_t))$, where we set the initial temperature $\tau=0.1$ and annealing rate $\gamma=0.001$. 
Monte-carlo estimate (with $S=100$ samples) of ELBO 
we optimized
w.r.t. the parameters $\lambda^b$ of a MDNF
using RMSprop with learning rate $0.001$.

The \textit{Asia} and \text{Sachs} have the same cardinality for all variables, but for \textit{Hepar II} the cardinality depends on the variable and ranges from 2 to 6. To handle the varying dimensionality in an environment designed for processing fixed-size tensors (in our case TensorFlow),
to represent $S$ $D$-dimensional samples, 
we use tensors of size $S \times D \times K$ 
with $K = \max\left(K_1, \dots, K_D\right)$, and map excess positions in one-hot encoded vectors (category positions with numbers larger than $K_d$) down to positions representing valid categories (by summing up zeros and ones from respective positions) only at the end -- when evaluating joint probability of observed and latent variables for a model. 
%
Entropy term can not be handled this way, but entropy for variables with $K$ categories bounds
the entropy for the original set of variables,
$H^K(x) \geq H^{K_1, \dots, K_d \dots K_D}(x)$,
and with \textit{Hepar II} we used the approximation.

\begin{figure*}[t]
\begin{center}
\includegraphics[width=0.30\textwidth]{figs/base_asia_B3_entropy.pdf}
\includegraphics[width=0.30\textwidth]{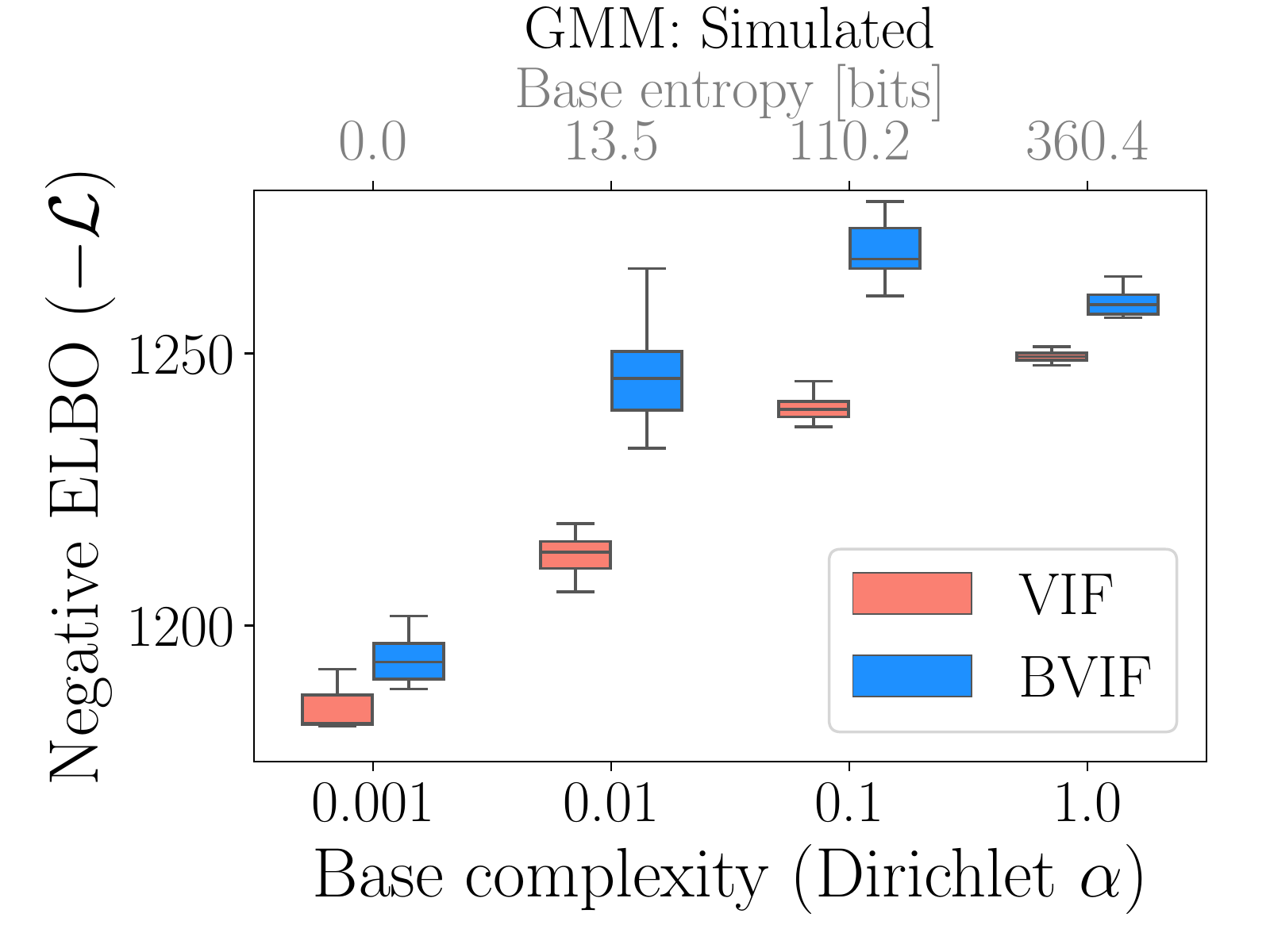}
\\
\includegraphics[width=0.28\textwidth]{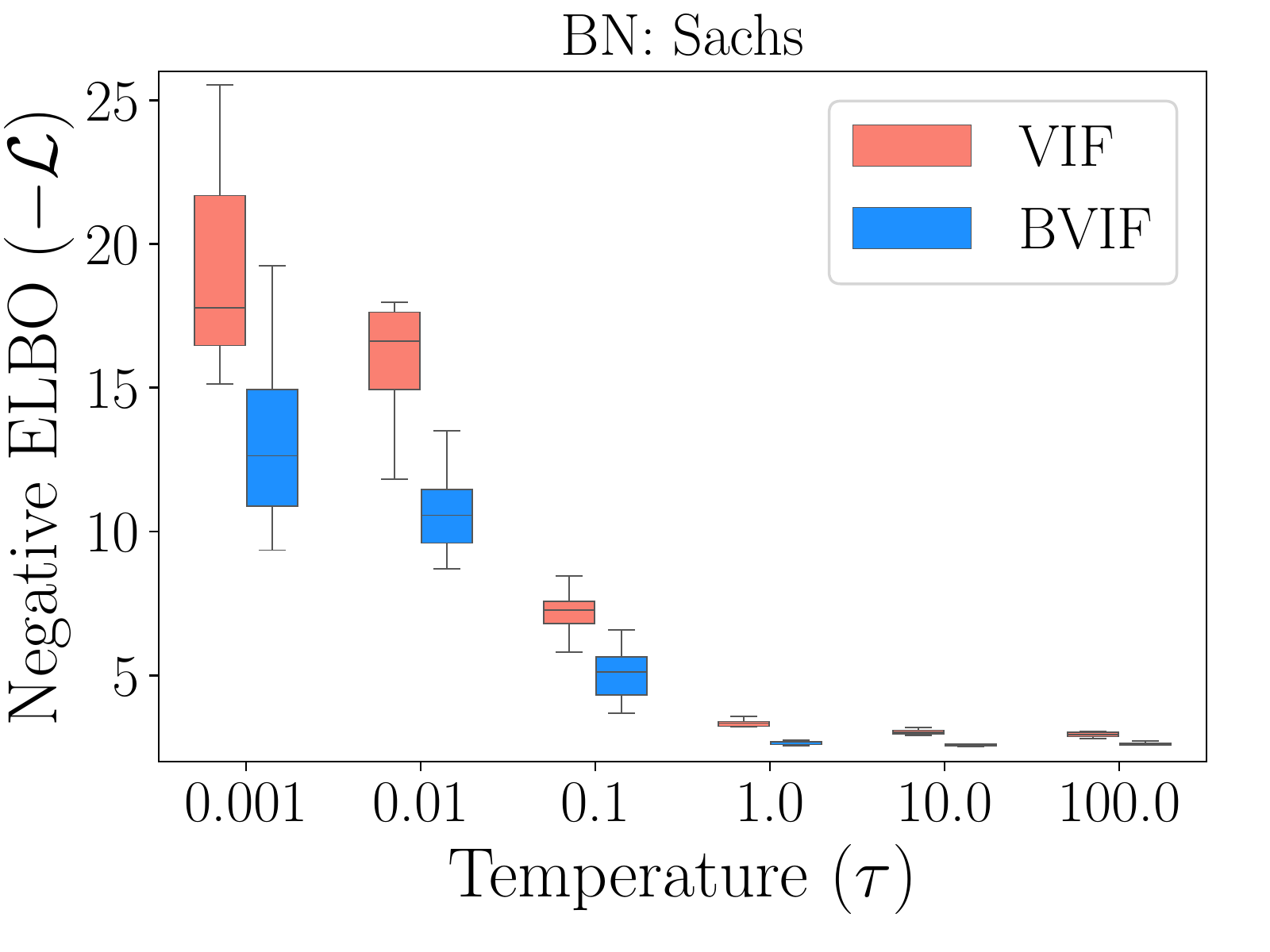}
\includegraphics[width=0.28\textwidth]{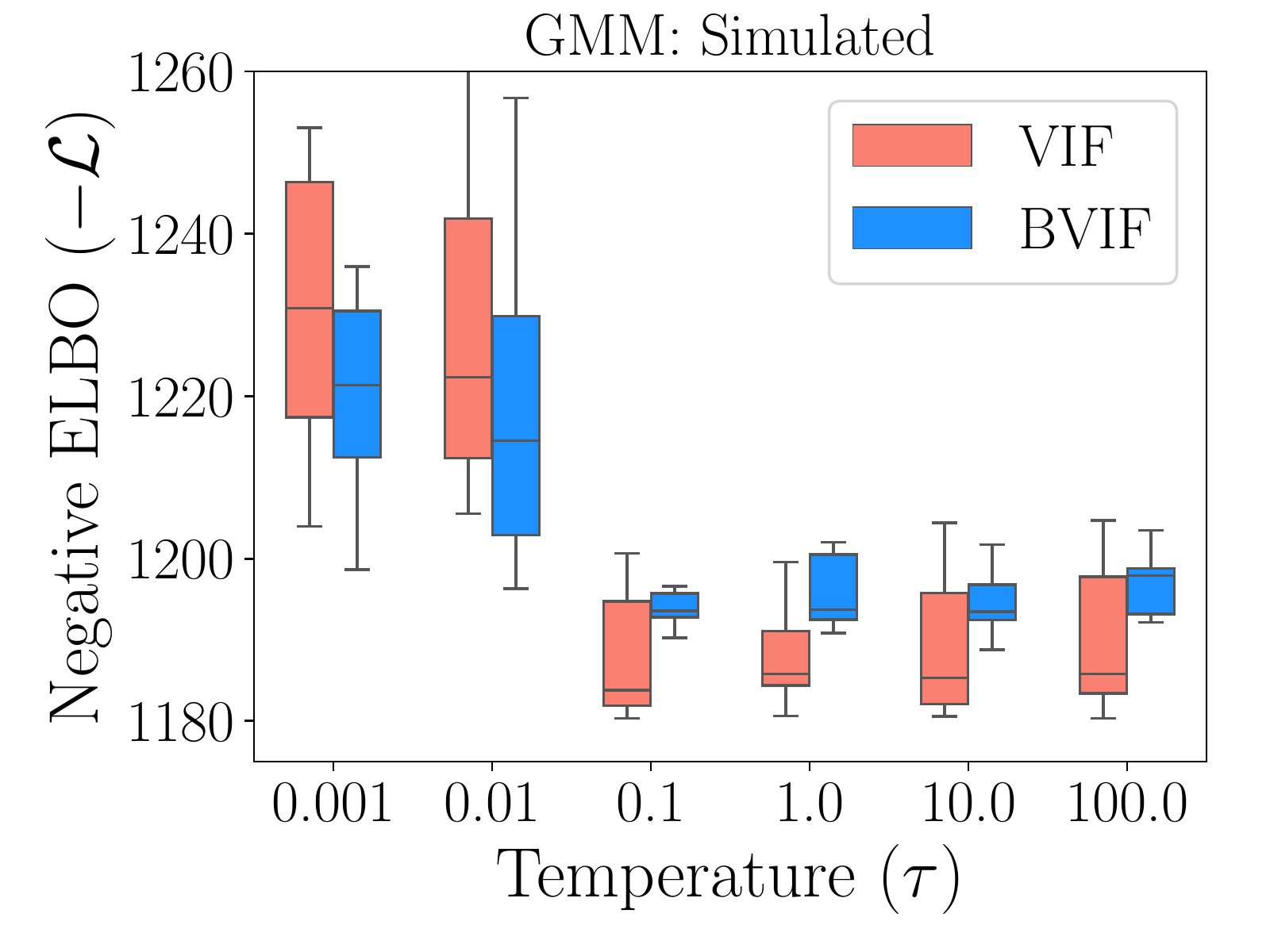}
\end{center}
\caption{Effects of base distribution (top row) and temperature $\tau$ (bottom row) on accuracy of posterior representation for BNs and GMMs (remaining hyperparameters set as previously).
}
\label{fig:base_entropy}
\label{fig:temperature_gmvi}
\end{figure*}

\paragraph{Additional Results}
To complement the analysis of BNs, we replicated the whole experiment on another model family. In particular, we provide an example of a model with both discrete and continuous latent variables, a Gaussian mixture model (GMM)~\citep{bishop2006pattern}.
GMM allocates $D$ observed data points $y_d \in \mathbb{R}^N$ to one of $K$ clusters with $D$ latent $K$-dimensional categorical variables $x_d$ (one per data point; the assignments are assumed to be conditionally independent). Note that we here denote -- somewhat unconventionally -- by $D$ the number of samples to emphasize that in our context the modeling task concerns learning the $D$-dimensional distribution of the latent allocations.

The model with multivariate normal component distributions is
\begin{align*}
 & p(\{y_d\}| \{x_d\}, \mu, \Lambda) = \prod_{d=1}^D \prod_{k=1}^K \mathcal{N}(y_d | \mu_k, \Lambda_k^{-1})^{x_{dk}}, \\
& p(\{x_d\} | \Pi) = \prod_{d=1}^D \prod_{k=1}^K \Pi_k^{x_dk},
\end{align*}
with priors 
\begin{align*}
& p(\Pi) = \text{Dir}(\Pi|\alpha_0), \\
& p(\mu, \Lambda) = \prod_{k=1}^K \mathcal{N}(\mu_k | \mu_0, (\beta_0 \Lambda_k)^{-1}) \mathcal{W}(\Lambda_k | W_0, \nu_0),
\end{align*}
where we used $\alpha_0=\frac{1}{K} \cdot \text{I}_{K \times K}$,
$\mu_{0n} = \frac{1}{D} \sum_{d=1}^{D} y_n$,
$\beta_0 = 1$,
$W_0 = \text{I}_{N \times N}$, and 
$\nu_0 = N$.

The model was trained using closed form Variational EM~\citep{bishop2006pattern} by alternating updates of the allocations (E-step) and the cluster parameters (M-step). Our implementation and choice of the hyperparameters follow a publicly available implementation\footnote{\smaller{\url{https://github.com/ctgk/PRML/blob/master/prml/rv/variational_gaussian_mixture.py}}}, where we replaced the closed-form E-step with stochastic gradient-based optimization of ELBO w.r.t. parameters of a MDNF modeling the distribution of latent allocations $x$. 
The optimization we performed using RMSprop optimizer (learning rate $0.1$) with $S = 100$ samples used for MC estimate of the objective. The posterior for $x$ factorizes and therefore we also used factorized flows with $\mu_d = ST(\text{softmax}(\lambda_d / \tau_t))$. That is, we passed the trainable parameters $\lambda$ directly through softmax and the straight-through (ST) operation. In variational EM with gradient-based E-step, stochastic optimization is performed multiple times, each time for slightly different clusters' found in M-step. The temperature annealing was adapted to this so that we anneal slightly in each step and then in each iteration of the step, e.g., in our schedule $t = \text{step}+\text{iteration}$ with initial temperature $\tau=10$ and
rate $\gamma=0.01$.

Figure~\ref{fig:gmmvi} compares BVIF and VIF for three data sets: 2-dimensional simulated data with 3 partially overlapping clusters (100 points each) with centers in (0,2),(1.7,-1),(-1.7,-1) and diagonal unit covariances; and for \textit{Google}\footnote{\url{https://archive.ics.uci.edu/ml/datasets/Tarvel+Review+Ratings}} (5456 data points with 24 features) and \textit{TripAdivsor}\footnote{\url{https://archive.ics.uci.edu/ml/datasets/Travel+Reviews}} (980 points with 10 features) travel reviews
\citep{renjith2018evaluation}. 
In all our experiments we used $K=3$ -- the same number of clusters that was used to generate the \textit{simulated} data, and that
was suggested for the \textit{Google} set by the authors~\citep{renjith2018evaluation} who provided the data. They also indicated $K=3$ is good, though not necessarily optimal, for \textit{TripAdivsor}.
We dropped BVI that performed very poorly for BNs, and again demonstrate that MDNF works as intended and sometimes improves on top of DNF (B=1), but for these simpler posteriors the difference is marginal because
in non-overlapping case all probability mass of a data point posterior is allocated to a single cluster.
Nevertheless, we see that MDNF can be used as plug-and-play approximation and works even when modeling a relatively large number of variables.

Finally, Figure~\ref{fig:base_entropy} presents additional illustrations regarding the choice of the base distribution and temperature hyperparameters, for both BNs and GMMs. We already showed in Figure~\ref{fig:bayesian_networks} (right) how delta distributions are ideal for BNs (repeated for convenience here; top left), and here we show that this is the case also for GMMs (top right). In addition, we show that the main result of MDNF being robust for the temperature choice (Section~\ref{sec:hyperparameter_experiment}) holds also in these experiments (bottom row); for both models and optimization algorithms large $\tau$ is good and there is no need to fine-tune the hyperparameter.

\begin{figure}[ht]
\begin{center}
\includegraphics[width=0.43\textwidth]{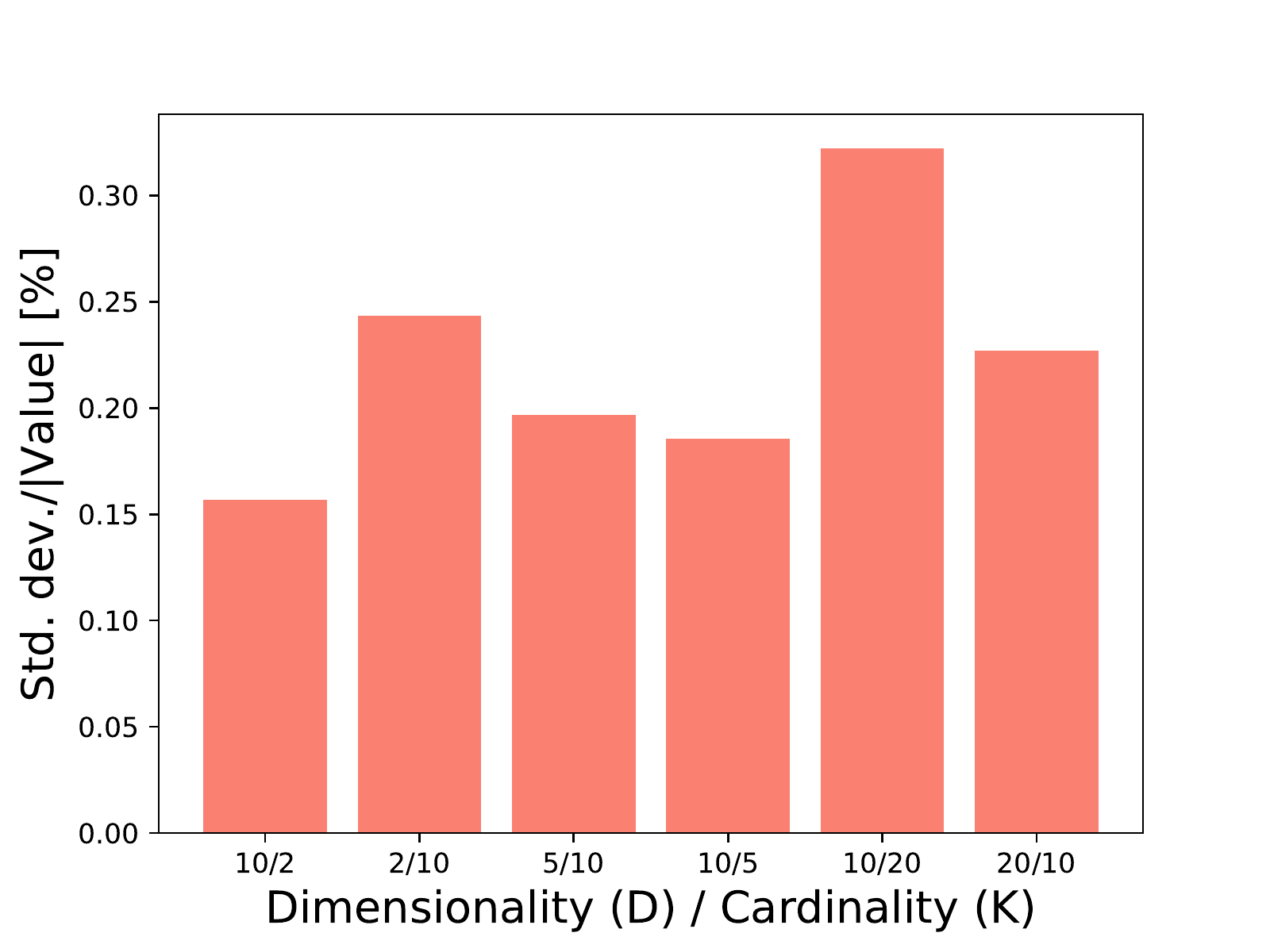}
\end{center}
\caption{Variance of ELBO estimator using MDNF on VAE (the worst case with $S=1$ sample, where $B=40$.) 
}
\label{fig:vae_elbo_variance}
\end{figure}

\subsection{Variance of the MDNF-based ELBO Estimator (Complements Section~4.3)}

In Section~\ref{sec:bias_and_variance} we stated that the variance of the variational objective estimate for MDNF is small. Figure~\ref{fig:vae_elbo_variance} verifies this empirically for a VAE model as used in Section~\ref{sec:vae_experiment}. We ran VIF with MDNF (B=40) for one full epoch, and then we sample the ELBO 100 times (always using only $S=1$ sample) and evaluate the empirical mean and variance of the estimate; the plot reports the variation as standard deviation normalized by the mean value. For all considered latent variable cardinalities, the deviation is less than 0.35\% (ratio of $0.0035$) of the mean estimate, despite using only a single sample to estimate the objective.

\end{document}